\documentclass[lettersize,journal]{IEEEtran}
\makeatletter
\DeclareRobustCommand\onedot{\futurelet\@let@token\@onedot}
\def\@onedot{\ifx\@let@token.\else.\null\fi\xspace}
\def\eg{\emph{e.g}\onedot} 
\def\ie{\emph{i.e}\onedot}

\makeatother

\usepackage{url}
\usepackage[utf8]{inputenc} 
\usepackage[T1]{fontenc}    
\usepackage{url}            
\usepackage{booktabs,tabularx,colortbl,multirow,multicol,array,makecell,tabularray}
\usepackage{amsfonts}       
\usepackage{nicefrac}       
\usepackage{microtype}      
\usepackage{xcolor}         
\usepackage[linesnumbered,ruled,vlined]{algorithm2e}
\usepackage{algpseudocode}
\usepackage{times}
\usepackage{epsfig}
\usepackage{epigraph}
\usepackage{graphicx}
\usepackage{amsmath}
\usepackage{amssymb}
\usepackage{float}
\usepackage{soul}
\usepackage{pifont}
\usepackage{wrapfig}
\usepackage{mathtools}
\usepackage{xspace}
\usepackage{enumitem}
\usepackage{threeparttable}

\usepackage{bm}
\usepackage{acronym}
\usepackage{xspace}
\usepackage{setspace}
\usepackage[dvipsnames,svgnames,x11names,table]{xcolor}
\usepackage[breaklinks,colorlinks,citecolor=blue]{hyperref}
\usepackage{orcidlink}
\usepackage{bbm}
\usepackage{siunitx}

\usepackage[capitalise]{cleveref}
\usepackage{overpic,wrapfig}
\usepackage[nocompress]{cite}

\usepackage[misc]{ifsym}
\usepackage{cite}
\usepackage{fontawesome}
\usepackage{svg}
\usepackage{float}
\usepackage{arydshln}
\usepackage{anyfontsize}
\usepackage{footnote}
\usepackage{bbm}
\usepackage{siunitx}
\usepackage{xcolor}
\usepackage{caption}
\usepackage{subcaption}
\usepackage{graphicx}
\usepackage{placeins}

\acrodef{vla}[VLA]{Vision-Language-Action}
\acrodef{vlg}[VLG]{Vision-Language-Grasp}
\acrodef{llm}[LLM]{Large Language Model}
\acrodef{vlm}[VLM]{Vision Language Model}
\acrodef{vae}[VAE]{variational autoencoder}
\acrodef{cvae}[CVAE]{conditional variational autoencoder}
\acrodef{rl}[RL]{Reinforcement Learning}
\acrodef{il}[IL]{Imitation Learning}
\acrodef{rope}[RoPE]{Rotary Position Encoding}
\acrodef{mlp}[MLP]{Multi-layer Perceptron}
\acrodef{ik}[IK]{inverse kinematics}
\acrodef{hoi}[HOI]{hand-object interaction}
\acrodef{cgr}[CGR]{Contact Grasp Representation}
\acrodef{dof}[DoF]{degree-of-freedom}
\acrodef{rag}[RAG]{Retrieval-Augmented Generation}
\acrodef{cot}[CoT]{Chain-of-Thought}
\acrodef{pcf}[PCF]{Principal Closing Frame}
\acrodef{camp}[CAMP]{Contact-Aware Multi-Pooling}
\acrodef{bce}[BCE]{Binary Cross-Entropy}
\acrodef{fps}[FPS]{Farthest Point Sampling}
\acrodef{sota}[SOTA]{state-of-the-art}
\acrodef{mot}[MoT]{Mixture-of-Transformer}
\acrodef{mgta}[MGTA]{Multiple Grasp Tracking Accuracy}
\acrodef{fn}[FN]{False Negatives}
\acrodef{fp}[FP]{False Positives}
\acrodef{idsw}[IDSW]{ID Switching}
\acrodef{mllm}[MLLM]{Multimodal Large Language Model}
\acrodef{cft}[CFT]{Contextual-Functional-Temporal}
\acrodef{sct}[SCT]{Spatial-Cognitive-Temporal}

\newcommand{\model}{AdaRoboVLG\xspace}

\newcommand{\redtext}[1]{\textcolor{red}{#1}}

\let\oldnl\nl
\newcommand{\nonl}{\renewcommand{\nl}{\let\nl\oldnl}}
\newcolumntype{x}{>{\columncolor{MistyRose}}c}
\newcolumntype{y}{>{\columncolor{LightCyan1}}c}
\newcolumntype{z}{>{\columncolor{LightOliveGreen}}c}

\definecolor{grasping}{RGB}{179,0,0}
\definecolor{placement}{RGB}{70,130,180}
\definecolor{goal_reaching}{RGB}{143,179,0}
\definecolor{gpurple}{RGB}{128, 0, 128}
\colorlet{LightOliveGreen}{OliveGreen!15}

\makeatletter
\renewcommand{\paragraph}{%
  \@startsection{paragraph}{4}%
  {\z@}{0ex \@plus 0ex \@minus 0ex}{0em}%
  {\hskip\parindent\normalfont\normalsize\bfseries}%
}
\makeatother

\crefname{algorithm}{Alg.}{Algs.}
\Crefname{algocf}{Algorithm}{Algorithms}
\crefname{section}{Sec.}{Secs.}
\Crefname{section}{Section}{Sections}
\crefname{table}{Tab.}{Tabs.}
\Crefname{table}{Table}{Tables}
\crefname{figure}{Fig.}{Figs.}
\Crefname{figure}{Figure}{Figures}
\crefname{equation}{Eq.}{Eqs.}
\Crefname{equation}{Equation}{Equations}
\crefname{appendix}{Appx.}{Appxs.}
\Crefname{appendix}{Appendix}{Appendices}

\newcommand{\refcolor}[1]{\textcolor{red}{#1}}
\crefformat{algorithm}{\refcolor{Alg.~(#2#1#3)}}
\Crefformat{algorithm}{\refcolor{Algorithm~(#2#1#3)}}

\crefformat{algocf}{\refcolor{Alg.~(#2#1#3)}}
\Crefformat{algocf}{\refcolor{Algorithm~(#2#1#3)}}

\crefformat{section}{\refcolor{Sec.~#2#1#3}}
\Crefformat{section}{\refcolor{Section~#2#1#3}}

\crefformat{table}{\refcolor{Tab.~#2#1#3}}
\Crefformat{table}{\refcolor{Table~#2#1#3}}

\crefformat{figure}{\refcolor{Fig.~#2#1#3}}
\Crefformat{figure}{\refcolor{Figure~#2#1#3}}

\crefformat{equation}{\refcolor{Eq.~(#2#1#3)}}
\Crefformat{equation}{\refcolor{Equation~(#2#1#3)}}

\crefformat{appendix}{\refcolor{Appx.~#2#1#3}}
\Crefformat{appendix}{\refcolor{Appendix~#2#1#3}}

\definecolor{gblue}{HTML}{4285F4}
\definecolor{gred}{HTML}{DB4437}
\definecolor{ggreen}{HTML}{0F9D58}

\newcommand{\transpose}{^{\mathsf{T}}}

\ifCLASSOPTIONcompsoc
  \usepackage[nocompress]{cite}
\else
  \usepackage{cite}
\fi
\ifCLASSINFOpdf
\else
\fi
\begin{document}

%

\title{Adaptive Vision-Language Grasping via Composable Foundation Priors and Generalizable Grasp Synthesis}

%
%
%
%

\author{
Sixu Yan\orcidlink{0009-0001-8502-1163},~\IEEEmembership{Student Member,~IEEE}, 
Shikang Wang\orcidlink{0009-0005-3314-8786},
Binhua Huang\orcidlink{0009-0003-8771-5729},
Xuanlai Tang\orcidlink{0009-0008-1340-019X},
Guohua Fan\orcidlink{0009-0004-6677-6845},
Fan Huang\orcidlink{0000-0002-6445-9553},
Haoxuan Li\orcidlink{0009-0006-5049-2359},
Yongkang Li\orcidlink{0009-0004-8684-2328},
Yuhan Li\orcidlink{0009-0008-8256-6281},
Bencheng Liao\orcidlink{0009-0009-6486-9235},
Zeyu Zhang\orcidlink{0000-0001-8929-134X},~\IEEEmembership{Member,~IEEE},
Wenyu Liu\orcidlink{0000-0002-4582-7488},~\IEEEmembership{Senior Member,~IEEE},
Hangxin Liu\orcidlink{0000-0002-3003-8611},~\IEEEmembership{Member,~IEEE}, and
Xinggang Wang\orcidlink{0000-0001-6732-7823},~\IEEEmembership{Senior Member,~IEEE}

\IEEEcompsocitemizethanks{
\IEEEcompsocthanksitem{Sixu Yan and Shikang Wang contributed equally to this work. \textit{(Corresponding authors: Xinggang Wang and Hangxin Liu.)}}
\IEEEcompsocthanksitem{Sixu Yan, Shikang Wang, Binhua Huang, Haoxuan Li, Yongkang Li, Yuhan Li, Wenyu Liu, and Xinggang Wang are with the School of Electronic Information and Communications, Huazhong University of Science and Technology, Wuhan 430074, China (e-mail:
\{yansixu, shikangwang, binhuahuang, lihaoxuan,
liyk, yuhanli, liuwy, xgwang\}@hust.edu.cn).}
\IEEEcompsocthanksitem{Xuanlai Tang is with KEENON Robotics Co., Ltd., Shanghai 201206, China (e-mail: tangxl@keenon.com).}
\IEEEcompsocthanksitem{Guohua Fan is with Suzhou Silicon Era Intelligent Technology Co., Ltd., Suzhou 215131, China (e-mail: m201272049@alumni.hust.edu.cn).}
\IEEEcompsocthanksitem{Fan Huang is with Suzhou Zhichuang Xinwei Technology Co., Ltd., Suzhou 215123, China, and also with the College of Materials, Xiamen University, Xiamen 361005, China (e-mail: phy.huangfan@gmail.com).}
\IEEEcompsocthanksitem{Bencheng Liao is with the School of Artificial Intelligence and Automation, Huazhong University of Science and Technology, Wuhan 430074, China (e-mail: bcliao@hust.edu.cn).}
\IEEEcompsocthanksitem{Zeyu Zhang is with the State Key Laboratory of General Artificial Intelligence, Beijing Institute for General Artificial Intelligence (BIGAI), Beijing 100080, China (e-mail: zhangzeyu@bigai.ai).}
\IEEEcompsocthanksitem{Sixu Yan is also with Hubei Automation Institute, Wuhan 430071, China.}
\IEEEcompsocthanksitem{Hangxin Liu is with the School of Artificial Intelligence, Peking University, Beijing 100871, China (e-mail: hx.liu@pku.edu.cn).}
}
}

\makeatletter
\g@addto@macro\@maketitle{
    \begin{figure}[H]
        \setlength{\linewidth}{\textwidth}
        \setlength{\hsize}{\textwidth}
        \centering
        \setcounter{figure}{0} 
        \includegraphics[width=\linewidth]{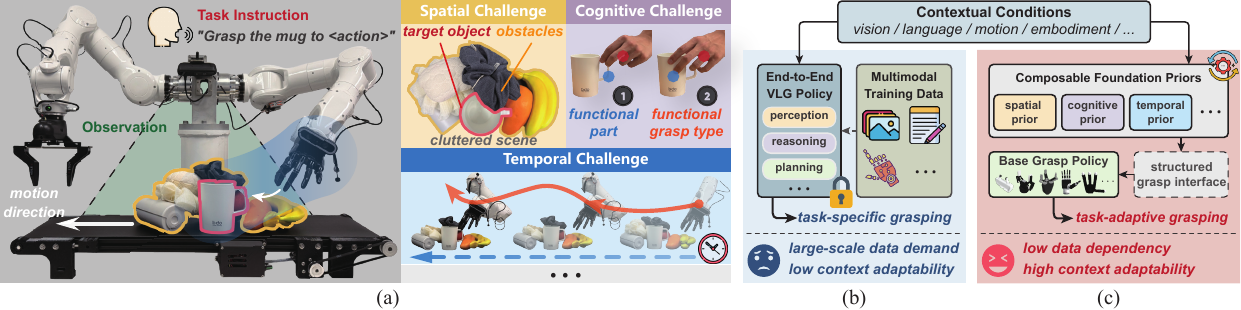}
        \vspace{-16pt}
        \caption{\textbf{Adaptive vision-language grasping.} (a) Real-world grasping occurs under diverse contextual conditions and task-dependent factors, each of which can substantially alter the desired grasp strategy. (b) End-to-end VLG policies typically entangle these factors within a single network trained on large-scale multimodal data, which limits their adaptability to unseen task contexts. In contrast, (c) \model decouples task-dependent understanding from physical grasp synthesis: foundation-model priors construct a structured grasp interface, from which a generalizable base policy synthesizes executable grasp poses.}
    \label{fig:teaser}
    \end{figure}
    \vspace{-26pt}
}
\makeatother  
\maketitle


\IEEEtitleabstractindextext{%
\begin{abstract}
This paper proposes \model, a task-adaptive \ac{vlg} framework that supports generalizable grasp synthesis across different robotic hands. Unlike existing \ac{vlg} methods that tightly couple foundation models with end-to-end grasp policies, \model learns an efficient generalizable base policy that generates and evaluates physically feasible grasp candidates through explicit kinematic mapping and force-closure-based stability estimation, while offloading task-dependent understanding to specialized foundation-model modules. These modules provide composable priors that are integrated into the grasp synthesis process, enabling contextually adaptive grasp synthesis without retraining the underlying grasp policy. Through extensive simulation and real-world experiments, we demonstrate that (i) the base policy exhibits efficient learning and strong cross-hand generalization, (ii) the framework effectively incorporates spatial, cognitive, and temporal priors to address three representative grasping challenges without compromising grasp synthesis performance compared to state-of-the-art methods, and (iii) these priors can operate jointly to enable functional grasping in cluttered and dynamic environments. These results indicate that decoupling physical grasp synthesis from task-dependent understanding provides a scalable paradigm for robotic grasping, allowing future advances in foundation models to be directly translated into improved grasp capabilities without redesigning or retraining the underlying grasp policy. Supplementary videos are available at \url{https://adarobovlg.github.io/}.
\end{abstract}


}

\IEEEdisplaynontitleabstractindextext

%
\IEEEpeerreviewmaketitle

\section{Introduction}

\label{sec:introduction}

\IEEEPARstart{G}{rasping} is one of the most fundamental skills required by robots for physical interaction~\cite{liu2024reconfigurable}. Despite substantial progress, the grasp generation problem remains challenging due to the difficulty in robustly selecting a grasp among almost infinite possible configurations between an end-effector and an object. More importantly, a grasp is inherently contextual. The same object may require entirely different grasp strategies depending on its surroundings, task objectives, and interaction dynamics. As illustrated in \cref{fig:teaser}\redtext{a}, executing the task instruction \textit{``Grasp the mug to $\langle$action$\rangle$''} oftentimes involves such intertwined factors. The target object may be partially occluded or covered by clutter, restricting feasible approach directions and grasp configurations. The grasp itself must satisfy functional requirements associated with subsequent object use: pour water or hand it over. And continuous adjustment is required for a successful grasp when the object or robot is moving. These spatial, cognitive, and temporal challenges frequently arise simultaneously, making robust and adaptive grasping particularly difficult.

Humans, in contrast, can effortlessly adapt their grasping behaviors to such combined challenges in varying situations. A key reason is that humans exploit rich priors accumulated from commonsense knowledge~\cite{zhu2020dark} and past experience~\cite{wilson2002six} to guide perception~\cite{goodale2013sight}, reasoning, and motor coordination~\cite{thelen1994dynamic}. Such priors enable efficient selection and adaptation of grasp strategies without requiring exhaustive experience for every possible situation. 

Analogous to the role of prior knowledge in human intelligence, foundation models have shown strong capabilities to provide informative priors for visual recognition and tracking~\cite{oquab2023dinov2,simeoni2025dinov3}, spatial understanding~\cite{chen2024ll3da,song2025robospatial}, embodied reasoning~\cite{RoboBrain1.0,RoboBrain2.0}, and robot control~\cite{black2024pi_0,intelligence2025pi_0.5}. In particular, their powerful visual recognition and language understanding capabilities have fueled the emergence of \ac{vlg} models~\cite{deng2025graspvla,zhong2025dexgraspvla,he2025dexvlg,beingbeyond2025beingh0,zhang2025omnidexvlg}, which exhibit strong generalization capabilities and support natural interactions by tightly integrating perception and grasp execution. However, existing \ac{vlg} methods predominantly incorporate foundation models within an end-to-end framework that directly maps visual observations and language instructions to grasp poses (see \cref{fig:teaser}\redtext{b}). Although effective, the performance of such specialized policies is largely constrained by the coverage of training data, model scale, and computational resources~\cite{deng2025graspvla,he2025dexvlg,zhang2025omnidexvlg}. As a result, scaling these policies to cover the vast diversity of grasp configurations, task intents, and environmental dynamics encountered in the real world remains impractical~\cite{choi2026autodex}.

Taking a different perspective, we propose \model, an adaptive vision-language grasping framework (see \cref{fig:teaser}\redtext{c}) that decouples task-dependent understanding from physical grasp synthesis. Specifically, we design a structured grasp interface consisting of object geometry, \acp{cgr}~\cite{sundermeyer2021contact}, and grasp types. A \ac{cgr} encodes a local contact primitive on the object surface, including the contact location, approach direction, closing direction, and grasp width. The grasp types follow the human grasp taxonomy~\cite{feix2015grasp} and are kinematically compatible with the target robotic hand. Centered on this structured grasp interface, we first learn an efficient base policy that converts the interface into hand-specific grasp candidates and selects the most stable candidate using a learned decision model that evaluates force-closure conditions~\cite{dai2015synthesis}. Throughout this process, complementary foundation-model priors for spatial understanding, cognitive reasoning, and temporal tracking construct or update the structured grasp interface, allowing the base policy to synthesize grasps that are both task-consistent and physically executable.

\begin{figure*}[t!]
    \centering
    \includegraphics[width=\linewidth]{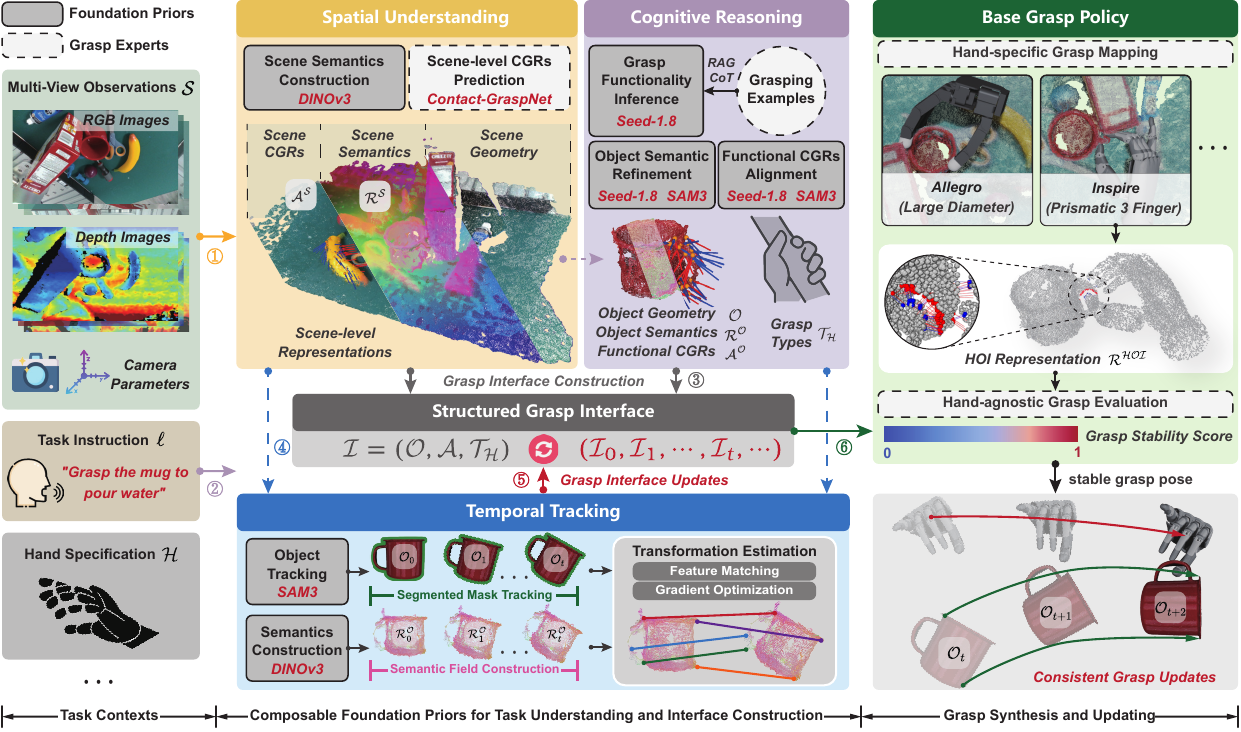}
    \caption{\textbf{Overview of \model.} To enable adaptive vision-language grasping, our framework explicitly decouples task-driven understanding from physical grasp synthesis. By integrating spatial, cognitive, and temporal priors, it constructs a structured grasp interface that directly guides an efficient base grasp policy to generate stable and executable grasp poses.}
    \label{fig:overview}
\end{figure*}

\cref{fig:overview} illustrates three representative examples. (i) By lifting DINOv3~\cite{simeoni2025dinov3} features extracted from multi-view RGB-D observations into 3D space, the spatial prior predicts global \acp{cgr}, providing spatially feasible contact constraints for grasp synthesis in cluttered scenes. (ii) By combining an \ac{llm} augmented with \ac{rag}~\cite{lewis2020retrieval} over a grasp-examples database with structured \ac{cot}~\cite{wei2022chain} reasoning, and SAM3~\cite{carion2025sam3}-based visual grounding, the cognitive prior localizes the target object, refines the \acp{cgr}, and infers appropriate grasp types for functional grasp generation according to language instructions. (iii) By combining the mask-tracking capability of SAM3 with the cross-frame feature consistency of DINOv3, the temporal prior estimates the motion of target geometry and \acp{cgr} online, enabling grasps to be continuously updated for dynamic objects. More importantly, these spatial, cognitive, and temporal priors modify the same structured grasp interface, allowing them to operate independently or jointly while enabling contextually adaptive grasp generation across a wide range of scenarios.

In simulation experiments, we first validate that the base policy is effective and generalizable, which not only exhibits efficient learning but also facilitates cross-hand deployment. Building on this base policy, we further evaluate the contextual adaptation of \model enabled by three types of example foundation priors. For clutter-aware grasping, the spatial prior enables \model to achieve an average success rate of 88.4\% across three robotic hands and three clutter levels on the DexGraspNet 2.0~\cite{zhang2024dexgraspnet} benchmark, achieving state-of-the-art results on the random and loose splits. For functional grasping, we evaluate the cognitive prior through two complementary experiments. In grasp functionality inference, it achieves functional-part and grasp-type accuracies of 94.0\% and 87.0\% on an open-set benchmark. In language-guided target grasping, \model achieves the best average success rates of 81.3\% and 86.0\% on simulation benchmarks built from GraspClutter6D~\cite{back2025graspclutter6d} and GraspNet-1Billion~\cite{fang2020graspnet}, respectively. Finally, for dynamic grasping, evaluations on GraspNet-1Billion show that the temporal prior enables perfect tracking association accuracy, the lowest translation errors, and competitive rotation errors compared to existing grasp-tracking methods.

In real-world settings, we further validate that the spatial, cognitive, and temporal priors can be jointly integrated to address more challenging grasping tasks. For functional grasping in cluttered scenes, \model achieves an overall success rate of 83.3\% over 510 trials across 102 everyday objects from six categories. On a moving conveyor belt, \model enables functional grasping in dynamic cluttered scenes, while performing online grasp updates at 5 Hz. We also evaluate the tracking robustness of \model under four types of human-induced disturbances, where it achieves an average success rate of 89.7\%.

The broader implications of this work extend beyond grasping performance itself. Decoupling physical grasp synthesis from task-dependent understanding not only improves data and training efficiency, but also establishes a stable interface through which increasingly capable foundation models can be integrated over time. Rather than repeatedly retraining end-to-end policies, future advances in perception and reasoning can be directly translated into improved grasping capabilities by updating only the corresponding foundation-prior modules.

\subsection{Contributions}
Our contributions are threefold:

\begin{enumerate}

\item We propose \model, an adaptive \ac{vlg} framework that decouples task-dependent understanding from physical grasp synthesis and bridges them through a structured grasp interface. It enables task adaptation, cross-hand grasp synthesis, and framework extensibility.

\item We develop an efficient base policy for generalizable grasp synthesis. It consists of an explicit kinematic mapping and a learned decision model that employs the proposed \ac{hoi} representation to encode local contact geometry relevant to force closure, supporting shared grasp stability evaluation across robotic hands.

\item We introduce composable spatial, cognitive, and temporal prior modules that construct and update the structured grasp interface for task-adaptive grasping under diverse and combined contextual conditions, such as cluttered scenes, functional requirements, and dynamic interactions.

\end{enumerate}

\subsection{Overview}
The remainder of this paper is organized as follows. \cref{sec:related_work} reviews the literature and compares existing research. \cref{sec:formulation} formulates adaptive vision-language grasping and defines the structured grasp interface. Building on this interface, \cref{sec:base_grasp_policy} develops the base policy for generalizable grasp synthesis, and \cref{sec:foundation_prior_instantiation} introduces the three composable foundation-prior modules for task-adaptive grasping under diverse contextual conditions. \cref{sec:sim_exp} and \cref{sec:real_exp} demonstrate the efficacy of \model through simulation and real-world experiments, respectively. Finally, \cref{sec:discussion_and_conclusion} concludes the paper with a discussion of limitations and future directions.

\section{Related Work}
\label{sec:related_work}

\subsection{Vision-Language-Conditioned Robotic Grasping} 
Vision-language grasping enables robots to perform precise grasps based on human instructions. To address the challenges illustrated in \cref{fig:teaser}\redtext{a}, recent works have advanced this direction by leveraging \acp{llm} and \acp{vlm} to (i) provide task-relevant guidance for grasp policies, such as semantic cues~\cite{tang2023graspgpt,yuan2024robopoint}, object affordances~\cite{tang2025affordgrasp,zhao2026affordance}, and spatial priors~\cite{ji2024graspsplats,zheng2024gaussiangrasper}; (ii) perform high-level task planning~\cite{qian2024thinkgrasp,liu2025robodexvlm} and functional reasoning~\cite{mirjalili2023lan,li2025Language}; and (iii) develop end-to-end \ac{vla} models~\cite{deng2025graspvla,he2025dexvlg,zhong2025dexgraspvla,zhang2025omnidexvlg}, trained on large-scale robotic datasets, to directly generate feasible grasp poses~\cite{he2025dexvlg,zhang2025omnidexvlg} or continuous actions~\cite{deng2025graspvla,zhong2025dexgraspvla}. 

However, these methods are often task- and hand-specific, requiring additional data collection and policy training when adapting to new tasks or embodiments. AdaRoboVLG addresses this limitation by decoupling task understanding from grasp synthesis through diverse foundation priors. New task requirements can be accommodated by updating the relevant priors, while new embodiments are supported by reusing the shared grasp policy, without modifying the entire framework.

\subsection{Learning-Based Multi-Finger Grasp Pose Synthesis} 
Learning-based multi-finger grasp synthesis can be broadly categorized into four representative paradigms. Specifically, hand-centric methods directly map perceptual observations to grasp actions~\cite{xu2021adagrasp,xu2023unidexgrasp,xu2023fast,wan2023unidexgrasp++,chen2025clutterdexgrasp}, enabling efficient inference but remaining constrained to specific hand embodiments. Object-centric methods improve cross-hand generalization using contact maps~\cite{varley2015generating,li2023gendexgrasp} or contact points~\cite{shao2020unigrasp,wu2025cedex}, but often require costly nonconvex optimization~\cite{wu2022learning}. Interaction-centric methods further balance generalization and efficiency by explicitly modeling hand-object relationships~\cite{she2022learning,wei2025dro,fei2025trograsp}. However, these paradigms primarily focus on object-level grasp synthesis. More recently, scene-centric methods have extended grasp synthesis to cluttered environments~\cite{li2022hgcnet,zhang2024dexgraspnet,zhang2026cadgrasp}, but their grasp generation and decision processes are typically coupled with specific hand embodiments.

Drawing from these efforts, \model decomposes grasp synthesis into two stages, with each stage built upon a hand-agnostic representation. During grasp generation, \acp{cgr}, which encode fundamental contact primitives, are converted into executable hand-specific grasp candidates through an explicit mapping function. During grasp decision, these candidates are encoded using the proposed \ac{hoi} representation and evaluated by a hand-agnostic decision model. This design retains efficient inference while supporting cross-hand grasp synthesis and extending naturally to complex multi-object scenes.

\subsection{Foundation Models in Embodied Intelligence}
Foundation models~\cite{firoozi2025foundation}, with their powerful capabilities in perception~\cite{simeoni2025dinov3,ravi2024sam2} and reasoning~\cite{achiam2023gpt,bai2025qwen2}, are becoming a vital cornerstone for general-purpose intelligent systems. They are widely used for high-level planning~\cite{RoboBrain1.0,RoboBrain2.0} and low-level control~\cite{black2024pi_0,intelligence2025pi_0.5} in robotics. Current applications can be broadly categorized into three types: (i) zero-shot utilization, where foundation priors are directly used for embodied navigation~\cite{li2025recogdrive,li2026unidrivevla} and manipulation~\cite{huang2023voxposer,huang2024rekep} without additional training; (ii) task-specific adaptation, where foundation models are adapted to robotic tasks through efficient finetuning~\cite{yuan2025roboengine,mei2025rogsam}; and (iii) integration into action policies, where foundation models are embedded into closed-loop perception-action pipelines, and some works further explore building policy foundation models~\cite{liu2024rdt,intelligence2025pi_0.5} trained on large-scale robotic datasets. 

Inspired by prior work, \model leverages and composes diverse foundation models to enable adaptive robotic grasping. Specifically, DINOv3~\cite{simeoni2025dinov3} and SAM3~\cite{carion2025sam3} are directly employed in a zero-shot manner for semantic understanding, target segmentation, and temporal tracking. Meanwhile, the \ac{llm} is augmented with grasp-relevant taxonomy knowledge to facilitate grasp functionality inference. The resulting spatial, cognitive, and temporal constraints are then unified into a structured grasp interface to guide physical grasp synthesis. By composing diverse foundation priors, \model can flexibly adapt to novel task contexts and grasping requirements.

\section{Problem Formulation and Grasp Interface}
\label{sec:formulation}

This section presents our formulation for adaptive vision-language grasping. We first introduce the essential notation and preliminaries in \cref{subsec:notation} and then define the problem and structured grasp interface in \cref{subsec:problem_formulation}.

\subsection{Notation and Preliminaries}
\label{subsec:notation}
Before delving into the formulation of \model, we start by defining the following mathematical notation:
\begin{itemize}[leftmargin=*,noitemsep,nolistsep]
    \item Bold lowercase letters represent vectors (\eg, $\boldsymbol{g}$), with the subscript $i$ denoting the $i$-th entry of the vector (\eg, $g_i$).
    \item Bold uppercase letters represent matrices (\eg, $\boldsymbol{T}$). The identity matrix of dimension $n$ is denoted by $\boldsymbol{I}_{n \times n}$. The superscript $\cdot^{\transpose}$ indicates the transpose.
    \item Calligraphic font letters are used to denote sets (\eg, $\mathcal{G}$). The notation $\lvert\cdot\rvert$ denotes the number of elements in a set.
\end{itemize}

\begin{figure}[t!]
    \centering
    \includegraphics[width=\linewidth]{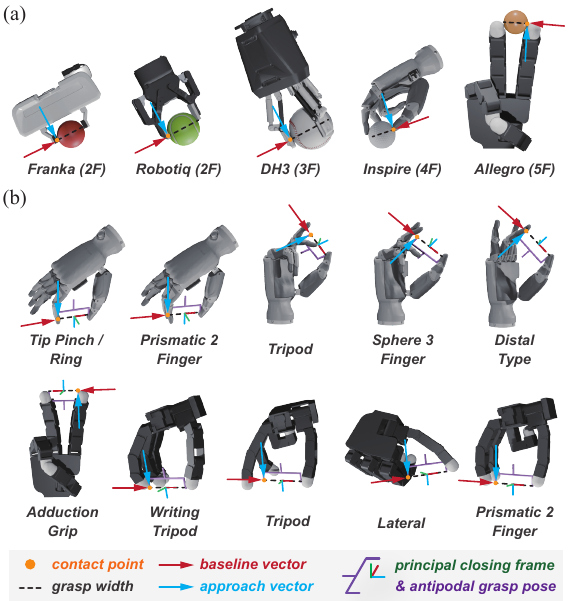}
    \caption{\textbf{Contact grasp representation.} This representation effectively captures a hand-agnostic contact primitive that can be (a) shared across diverse end-effectors, ranging from parallel grippers to multi-finger robotic hands, and (b) adaptable to various functional grasp types.}
    \label{fig:cgr}
\end{figure}

\subsection{Problem Formulation and Structured Grasp Interface}
\label{subsec:problem_formulation}
A multi-finger grasp pose $\boldsymbol{g}$ is formally defined as:
\begin{equation}
    \label{eq:dex_grasp_pose}
    \begin{aligned}
    \boldsymbol{g} = \left( \boldsymbol{T}, \boldsymbol{q} \right), \; \boldsymbol{T}=\left[
        \begin{array}{cccc}
            & \boldsymbol{R} & & \boldsymbol{t} \\ 0 & 0 & 0 & 1
        \end{array}
    \right],
    \end{aligned}
\end{equation}
where $\boldsymbol{T} \in \mathrm{SE}(3)$ denotes the hand base pose in the world frame, comprising a rotation matrix $\boldsymbol{R} \in \mathbb{R}^{3 \times 3}$ and a translation vector $\boldsymbol{t} \in \mathbb{R}^{3}$. $\boldsymbol{q} \in \mathbb{R}^{n}$ characterizes the joint configuration of an $n$-\ac{dof} multi-finger robotic hand. The goal of adaptive vision-language grasping is to synthesize a set of executable grasp poses $\mathcal{G}=\{\boldsymbol{g}_{i}\}_{i=1}^{|\mathcal{G}|}$ that are compatible with hand embodiment $\mathcal{H}$ and satisfy the task context $\mathcal{C}$.

Instead of learning an end-to-end network that maps the task context to grasp poses, we decompose adaptive vision-language grasping into task-dependent understanding and physical grasp synthesis, and bridge them through a structured grasp interface. Under this formulation, the task context $\mathcal{C}$ is first converted into a structured grasp interface $\mathcal{I}$. Subsequently, executable grasp poses $\mathcal{G}^{*}$ are synthesized from this interface:
\begin{equation}
\label{eq:conditioned_base_policy}
    \begin{aligned}
    f_{\tau}:\mathcal{C}\rightarrow\mathcal{I}, \;
    f_{\pi}:\mathcal{I}\times\mathcal{H}\rightarrow\mathcal{G}^{*}.
    \end{aligned}
\end{equation}
Formally, we define this interface as
\begin{equation}
    \label{eq:structured_grasp_interface}
    \begin{aligned}
    \mathcal{I}
    =
    \left(
    \mathcal{O},
    \mathcal{A},
    \mathcal{T}_{\mathcal{H}}
    \right),
    \end{aligned}
\end{equation}
where $\mathcal{O}$ denotes the object geometry, $\mathcal{A}=\left\{\boldsymbol{a}_{j}\right\}_{j=1}^{|\mathcal{A}|}$ denotes the set of \acp{cgr} that characterize the fundamental contact primitives for stable grasping, and $\mathcal{T}_{\mathcal{H}}=\{\tau_k\}_{k=1}^{|\mathcal{T}_{\mathcal{H}}|}$ represents the set of grasp types that follow the human grasp taxonomy and are kinematically compatible with the hand embodiment $\mathcal{H}$. Here, $f_{\pi}(\cdot)$ denotes the base grasp policy, which synthesizes executable grasp poses from the structured interface $\mathcal{I}$ and the hand embodiment $\mathcal{H}$. $f_{\tau}(\cdot)$ denotes the transformation from task context $\mathcal{C}$ to the interface $\mathcal{I}$, which can be instantiated with diverse composable foundation priors.
\section{Base Grasp Policy}
\label{sec:base_grasp_policy}

This section presents the implementation details of the base grasp policy $f_{\pi}(\cdot)$. We sequentially describe the policy design principle (\cref{subsec:base_policy_design_principle}), the \ac{cgr} definition (\cref{subsec:base_policy_cgr_representation}), the hand-specific grasp mapping (\cref{subsec:base_policy_hand_mapping}), the \ac{hoi} representation (\cref{subsec:base_policy_hoi_representation}), and the hand-agnostic grasp evaluation (\cref{subsec:base_policy_hand_agnostic_decision}).

\begin{figure}[t!]
    \centering
    \includegraphics[width=\linewidth]{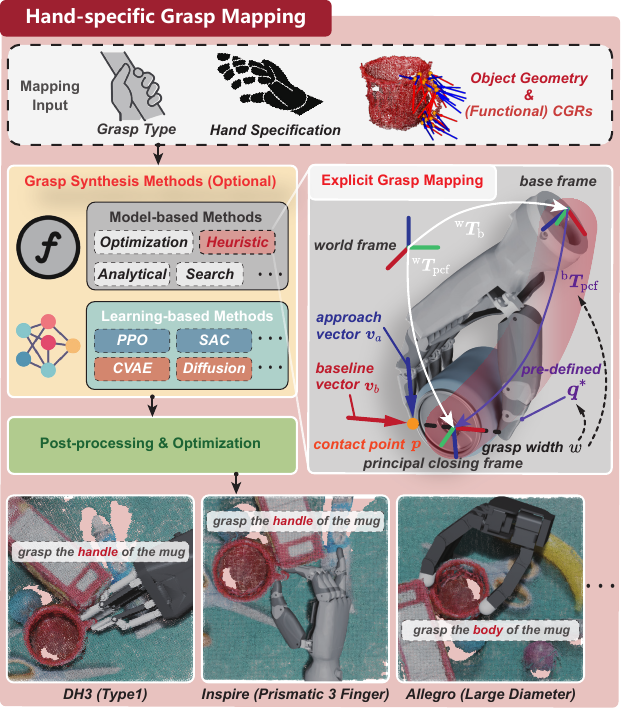}
    \caption{\textbf{Hand-specific grasp mapping.} This function maps the object geometry, \acp{cgr}, and grasp type into taxonomy-consistent grasp poses for a specific robotic hand.}
    \label{fig:grasp_mapping}
\end{figure}

\subsection{Design Principle}
\label{subsec:base_policy_design_principle}

Given the grasp interface $\mathcal{I}$ defined in \cref{eq:structured_grasp_interface}, the base policy converts it into executable grasp poses for a target robotic hand. To support efficient and cross-hand grasp synthesis, the policy adopts a two-stage process consisting of hand-specific grasp mapping and hand-agnostic grasp evaluation.

For grasp mapping, the interface $\mathcal{I}$ is mapped to candidate grasp poses for a specific robotic hand $\mathcal{H}$. Specifically, a hand-specific mapping function $\mathcal{M}_{\mathcal{H}}(\cdot)$ takes the target object geometry $\mathcal{O}$, a \ac{cgr} descriptor $\boldsymbol{a}_{j}$, and a grasp type $\tau_{k}$ as input, and generates the target grasp poses $\mathcal{G}_{j,k}$:
\begin{equation}
    \label{eq:hand_centric_grasp_mapping}
    \begin{aligned}
    \mathcal{G}_{j,k}
    =
    \mathcal{M}_{\mathcal{H}}
    \left(
        \mathcal{O},
    \boldsymbol{a}_{j},
    \tau_k
    \right),
    \end{aligned}
\end{equation}
where $\mathcal{G}_{j,k}=\{\boldsymbol{g}_{j,k}^{(i)}\}_{i=1}^{|\mathcal{G}_{j,k}|}$ is the set of grasp pose candidates. The complete set of grasp candidates for $\boldsymbol{a}_{j}$ is computed by aggregating candidates across all such grasp types:
\begin{equation}
    \label{eq:all_grasp_poses}
    \begin{aligned}
    \mathcal{G}_{j} = \bigcup_{k=1}^{|\mathcal{T}_{\mathcal{H}}|} \mathcal{G}_{j,k} = \bigcup_{\tau_k \in \mathcal{T}_{\mathcal{H}}} \mathcal{M}_{\mathcal{H}}(\mathcal{O}, \boldsymbol{a}_{j}, \tau_k).
    \end{aligned}
\end{equation}

Subsequently, we estimate the quality of each grasp candidate. We first extract the \ac{hoi} representation $\mathcal{R}^{\mathcal{HOI}}$ from the hand-object contact interface relevant to force closure~\cite{dai2015synthesis}. This process is achieved by an extractor $\mathcal{E}(\cdot)$, which takes the object geometry $\mathcal{O}$, the hand specification $\mathcal{H}$, and a grasp pose $\boldsymbol{g}_{j,k}^{(i)}$ as input, and produces the \ac{hoi} representation $\mathcal{R}^{\mathcal{HOI}}$:
\begin{equation}
    \label{eq:interaction_extract}
    \mathcal{R}^{\mathcal{HOI}} = \mathcal{E} \left(\mathcal{O},\mathcal{H},\boldsymbol{g}_{j,k}^{(i)}\right).
\end{equation}
Then, a hand-agnostic grasp decision model $\mathcal{D}(\cdot)$ is introduced to predict the grasp success probability $\beta$, conditioned on the \ac{hoi} representation $\mathcal{R}^{\mathcal{HOI}}$:
\begin{equation}
    \label{eq:grasp_decision_model}
    \begin{aligned}
    \beta = \mathcal{D}(\mathcal{R}^{\mathcal{HOI}}).
    \end{aligned}
\end{equation}

The objective of the base policy is to find a set of top-$K$ grasp poses $\mathcal{G}^{*}$ that maximizes the predicted success rate:
\begin{equation}
    \label{eq:grasp_optimization_objective}
    \mathcal{G}^{*}=\underset{\mathcal{G} \subset \bigcup_j \mathcal{G}_j,|\mathcal{G}|=K}{\arg \max } \mathbb{E}_{\boldsymbol{g}_{j,k}^{(i)}}\bigg[\mathcal{D}\left(\mathcal{E}(\mathcal{O}, \mathcal{H}, \boldsymbol{g}_{j,k}^{(i)})\right)\bigg],
\end{equation}
where $\boldsymbol{g}_{j,k}^{(i)} \in \mathcal{M}_{\mathcal{H}}(\boldsymbol{a}_j, \tau_k) \cap \mathcal{G}$, $\boldsymbol{a}_j \in \mathcal{A}$, and $\tau_k \in \mathcal{T}_{\mathcal{H}}$.

\subsection{Contact Grasp Representation}
\label{subsec:base_policy_cgr_representation}

The \ac{cgr} serves as a fundamental descriptor of graspability, encoding the inherent, local information of the object surface to establish stable grasping contacts. Following prior work~\cite{sundermeyer2021contact}, we parameterize each \ac{cgr} $\boldsymbol{a}$ as:
\begin{equation}
    \label{eq:base_cgr_definition}
    \begin{aligned}
    \boldsymbol{a}
    =
    \left(
    \boldsymbol{p}_{c},
    \boldsymbol{v}_{a},
    \boldsymbol{v}_{b},
    p,
    w
    \right),
    \end{aligned}
\end{equation}
where $\boldsymbol{p}_{c}\in\mathbb{R}^{3}$ denotes the potential contact point on the object surface, $p \in [0, 1]$ denotes the probability that this point is a stable contact point, $\boldsymbol{v}_a \in \mathbb{R}^3, \|\boldsymbol{v}_a\|=1$ denotes the grasp approach vector, $\boldsymbol{v}_b \in \mathbb{R}^3, \|\boldsymbol{v}_b\|=1$ denotes the grasp baseline (closing) vector, and $w \in \mathbb{R}^{+}$ denotes the grasp width. As shown in \cref{fig:cgr}, this representation captures fundamental, hand-agnostic grasp contact primitives that can be shared across different end-effectors and grasp types.

\begin{figure*}[t!]
    \centering
    \includegraphics[width=\linewidth]{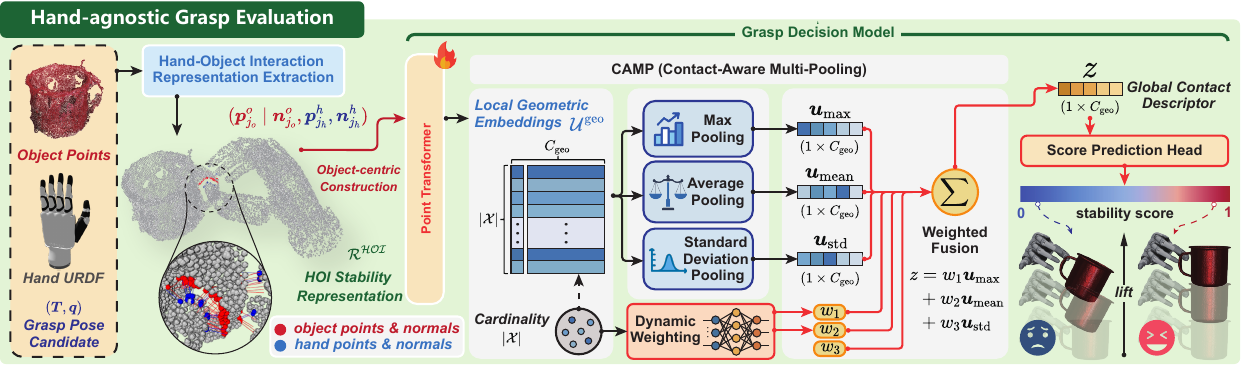}
    \caption{\textbf{Hand-agnostic grasp evaluation.} This module evaluates the physical stability of synthesized grasp candidates in a strictly hand-agnostic manner. To achieve this, it first extracts an \ac{hoi} representation to explicitly capture the local contact geometry. Leveraging a PointTransformer~\cite{wu2024ptv3} backbone and a contact-aware feature extraction module, the grasp decision model dynamically aggregates these geometric features to predict a continuous stability score for each grasp pose.}
    \label{fig:grasp_evaluation}
\end{figure*}

\subsection{Hand-Specific Grasp Mapping}
\label{subsec:base_policy_hand_mapping}

This subsection instantiates the mapping function $\mathcal{M}_{\mathcal{H}}(\cdot)$ introduced in \cref{eq:hand_centric_grasp_mapping}. In principle, $\mathcal{M}_{\mathcal{H}}(\cdot)$ can be implemented by model-based~\cite{liu2021synthesizing,meattini2023human} or learning-based~\cite{chen2025dexonomy,li2025Language,zhang2025omnidexvlg} grasp synthesis methods. To achieve efficient grasp synthesis, we instantiate it as an explicit mapping function, following the established methodologies in prior works~\cite{yan2024dexterous,yan2024surprisingly,fang2025anydexgrasp}.

For any given grasp type, the hand closing synergy (the first principal component of all kinematically feasible joint configurations) captures approximately 50\%--70\% of the kinematic variance~\cite{santello1998postural,mason2001hand,jarque2016using}. Building upon this insight, we instantiate the feasible kinematics for each grasp type as a 1-D parametric manifold within the hand configuration space $\mathcal{Q}_{\mathcal{H}} \subset \mathbb{R}^n$, parameterized solely by the grasp width $w$. For efficient inference, we precompute this manifold as a hand-specific lookup table. Formally, it is defined as
\begin{equation}
    \label{eq:base_hand_centric_mapping_instantiation}
    \begin{aligned}
    \mathcal{L}_{\mathcal{H}}:
    \mathcal{T}_{\mathcal{H}}
    \times
    \mathbb{R}^{+}
    \rightarrow
    \mathcal{Q}_{\mathcal{H}}
    \times
    \mathrm{SE}(3),
    \end{aligned}
\end{equation}
which maps a grasp type $\tau$ and a grasp width $w$ to a hand configuration $\boldsymbol{q}^{*}$ and the corresponding \ac{pcf} transformation ${}^{\mathrm{b}}\boldsymbol{T}_{\mathrm{pcf}}$ that denotes the pose matrix of the \ac{pcf} with respect to the hand base frame, as shown in \cref{fig:grasp_mapping}.

To construct this lookup table offline, we discretize the grasp width into a sequence $\{w_l\}$ and solve for the joint configuration $\boldsymbol{q}_{l,\tau}^{*}$ of each grasp type $\tau$:
\begin{equation}
    \label{eq:base_solving_predefined_config}
    \begin{aligned}
    \boldsymbol{q}_{l,\tau}^{*}
    =
    \underset{\boldsymbol{q}}{\arg\min}\;
    &
    \left\|
    \mathcal{J}_{w}(\boldsymbol{q},\mathcal{P}_{c})-w_l
    \right\|_2^2
    +
    \gamma
    \left\|
    \boldsymbol{q}-\boldsymbol{q}_{l-1,\tau}^{*}
    \right\|_2^2, \\
    \textit{s.t.}\;&
    \boldsymbol{q}_{\min}
    \leq
    \boldsymbol{q}
    \leq
    \boldsymbol{q}_{\max},
    \;
    \mathbf{g}_{\tau}(\boldsymbol{q})=\mathbf{0}.
    \end{aligned}
\end{equation}
The recursion is initialized with $\boldsymbol{q}_{0,\tau}^*=\boldsymbol{q}_{\text{init},\tau}$. Here, $\mathcal{J}_w(\boldsymbol{q}, \mathcal{P}_{c})$ computes the grasp width from a virtual plane constructed by transforming a set of predefined contact points $\mathcal{P}_{c}$ through forward kinematics. In \cref{eq:base_solving_predefined_config}, the first term minimizes the deviation from the target width, while the second term regularizes the solution toward the previous discretized configuration to ensure kinematic smoothness. The constraint $\mathbf{g}_{\tau}(\boldsymbol{q})=\mathbf{0}$ enforces taxonomy-specific kinematic constraints for grasp type $\tau$. After solving $\boldsymbol{q}_{l,\tau}^*$, we analytically compute the corresponding \ac{pcf} transformation ${}^{\text{b}}\boldsymbol{T}_{\text{pcf},l,\tau}$ from the same contact plane. Finally, all pairs $(w_l, \boldsymbol{q}_{l,\tau}^*, {}^{\text{b}}\boldsymbol{T}_{\text{pcf},l,\tau})$ are precomputed and stored as a lookup table for efficient inference.

During inference, we can reconstruct the world-frame \ac{pcf} from a \ac{cgr} $\boldsymbol{a}_{j}=(\boldsymbol{p}_{c},\boldsymbol{v}_{a},\boldsymbol{v}_{b},p,w)_{j}$:
\begin{equation}
    \label{eq:base_world_pcf}
    \renewcommand{\arraystretch}{1.25}
    {}^{\mathrm{w}}\boldsymbol{T}_{\mathrm{pcf},j}
    =
    \left[
    \begin{array}{ccc|c}
        \boldsymbol{v}_{b} &
        \boldsymbol{v}_{a}\times\boldsymbol{v}_{b} &
        \boldsymbol{v}_{a} &
        \boldsymbol{p}_{c}+\displaystyle\frac{w}{2}\boldsymbol{v}_{b} \\
        0 & 0 & 0 & 1
    \end{array}
    \right]_{j}.
\end{equation}
We retrieve the hand configuration $\boldsymbol{q}^{*}_{k,\tau_k}$ and hand-frame \ac{pcf} transformation ${}^{\mathrm{b}}\boldsymbol{T}_{\mathrm{pcf},j,\tau_k}$ according to $w_j$ and $\tau_k$. The hand base pose and joint configuration are then computed as
\begin{equation}
    \label{eq:base_hand_pose_from_pcf}
    \begin{aligned}
    \boldsymbol{T}_{j,k}
    =
    {}^{\mathrm{w}}\boldsymbol{T}_{\mathrm{pcf},j}
    \left(
    {}^{\mathrm{b}}\boldsymbol{T}_{\mathrm{pcf},j,\tau_k}
    \right)^{-1},
    \;
    \boldsymbol{q}_{j,k}
    =
    \boldsymbol{q}^{*}_{j,\tau_k}.
    \end{aligned}
\end{equation}
This produces the candidate grasp $\boldsymbol{g}_{j,k}=(\boldsymbol{T}_{j,k},\boldsymbol{q}_{j,k})$. We subsequently augment each candidate by sampling five discrete depths (\qtyrange[range-phrase=--]{0}{0.05}{\meter}) along the approach direction $\boldsymbol{v}_{a,k}$. Finally, physically infeasible poses are filtered out via a fast sphere-based collision check using the Foam~\cite{coumar2025foam} library, yielding the final set of grasp candidates $\mathcal{G}_{j,k} = \left\{ \boldsymbol{g}_{j,k}^{(i)} \right\}$.

\begin{figure*}[t!]
    \centering
    \includegraphics[width=\linewidth]{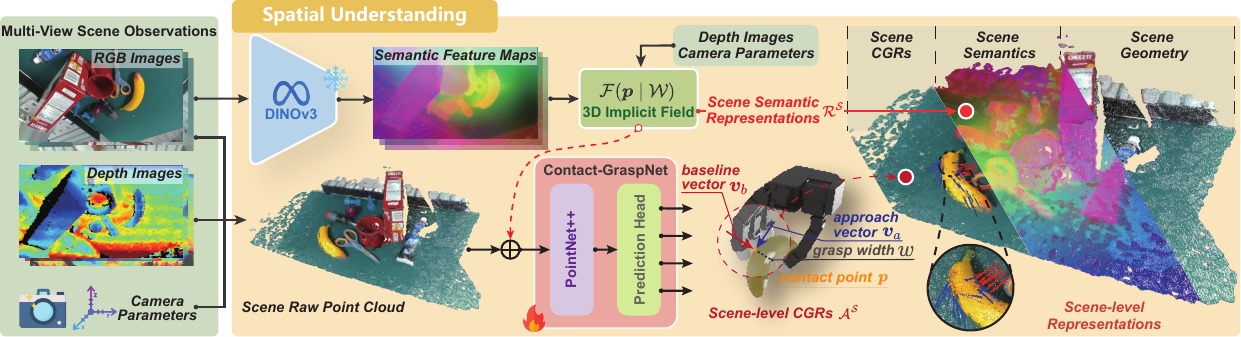}
    \caption{\textbf{Spatial prior module.} Given multi-view RGB-D observations, the spatial prior module leverages the DINOv3~\cite{simeoni2025dinov3} features to construct scene semantic representations $\mathcal{R}^{\mathcal{S}}$ and predict scene-level \acp{cgr} $\mathcal{A}^{\mathcal{S}}$. The predicted \acp{cgr} provide spatially feasible contact candidates for the base grasp policy to perform clutter-aware grasp planning.}
    \label{fig:spatial_understanding}
\end{figure*}

\subsection{Hand-Object Interaction Representation}
\label{subsec:base_policy_hoi_representation}

A grasp is considered stable if it can resist arbitrary external wrenches. This state is defined by the \textit{force-closure condition}, which must satisfy the following criteria~\cite{dai2015synthesis}:
\begin{equation}
    \label{eq:base_force_closure_condition}
    \begin{aligned}
    &
    \boldsymbol{G}\boldsymbol{f}=0, \\
    &
    \boldsymbol{G}\boldsymbol{G}^{\transpose}
    >
    \epsilon
    \boldsymbol{I}_{6\times6}, \\
    &
    \boldsymbol{f}_{i}^{\transpose}
    \boldsymbol{n}_{i}
    >
    \frac{1}{\sqrt{\mu^{2}+1}}
    \left|
    \boldsymbol{f}_{i}
    \right|,
    \end{aligned}
\end{equation}
where $\mu$ is the friction coefficient, $\boldsymbol{f}$ is the vector of contact forces acting at all contact points, $\boldsymbol{G}$ is the grasp matrix determined by the positions of the contact points, and $\boldsymbol{f}_i$ and $\boldsymbol{n}_i$ are the contact force and the surface normal at the $i$-th contact point. From visual perception, the information we can extract is the positions and normals of potential contact points.

Consequently, we define the \ac{hoi} representation $\mathcal{R}^{\mathcal{HOI}}$ as the set of coordinates and normals of the contact points at the hand-object interface. Specifically, given a grasp candidate $\boldsymbol{g}_{j,k}^{(i)}$ and the hand specification, we first extract the hand points $\mathcal{P}_{\mathcal{H}}=\{\boldsymbol{p}_{n_h}^h\}_{n_h=1}^{N_h}$ and normals $\mathcal{N}_{\mathcal{H}}=\{\boldsymbol{n}_{n_h}^h\}_{n_h=1}^{N_h}$ via forward kinematics. Analogously, the object points and normals are denoted as $\mathcal{P}_{\mathcal{O}}=\left\{\boldsymbol{p}_{n_o}^o\right\}{_{n_o=1}^{N_o}}$ and $\mathcal{N}_{\mathcal{O}}=\left\{\boldsymbol{n}_{n_o}^o\right\}{_{n_o=1}^{N_o}}$. To identify contact regions, we compute the Euclidean distance from each object point to the hand surface:
\begin{equation}
    \label{eq:base_hand_object_distance}
    d_{n_o}
    =
    \min_{n_h}
    \left\|
    \boldsymbol{p}_{n_o}^{o}
    -
    \boldsymbol{p}_{n_h}^{h}
    \right\|_2.
\end{equation}
Points whose distance is below a proximity threshold $\delta_d$ are treated as contact pairs. The valid contact-pair index set is
\begin{equation}
    \label{eq:base_contact_pair_extraction}
    \begin{aligned}
    \mathcal{X}
    =
    \left\{
    (n_o,n_h^{*})
    \mid
    d_{n_o}<\delta_d,\;
    n_h^{*}
    =
    \underset{n_h}{\arg\min}
    \left\|
    \boldsymbol{p}_{n_o}^{o}
    -
    \boldsymbol{p}_{n_h}^{h}
    \right\|_2
    \right\}.
    \end{aligned}
\end{equation}
The \ac{hoi} representation is then constructed as
\begin{equation}
    \label{eq:base_hoi_representation_definition}
    \begin{aligned}
    \mathcal{R}^{\mathcal{HOI}}
    =
    \left\{
    \left(
    \boldsymbol{p}_{n_o}^{o},
    \boldsymbol{n}_{n_o}^{o},
    \boldsymbol{p}_{n_h}^{h},
    \boldsymbol{n}_{n_h}^{h}
    \right)
    \mid
    (n_o,n_h)\in\mathcal{X}
    \right\}.
    \end{aligned}
\end{equation}
This representation provides a unified evaluation interface across different hand embodiments.

\subsection{Hand-Agnostic Grasp Evaluation}
\label{subsec:base_policy_hand_agnostic_decision}

This subsection instantiates the grasp decision model $\mathcal{D}(\cdot)$ in \cref{eq:grasp_decision_model}. We implement $\mathcal{D}(\cdot)$ as a lightweight point-based neural network that predicts a scalar stability score from $\mathcal{R}^{\mathcal{HOI}}$.

As illustrated in \cref{fig:grasp_evaluation}, we adopt an object-centric formulation to construct the \ac{hoi} representation. For each valid contact pair $(n_o, n_h)$, the object contact point $\boldsymbol{p}_{n_o}^o \in \mathbb{R}^3$ serves as the spatial coordinate, while the remaining geometric attributes form the point-wise feature vector $[\boldsymbol{n}_{n_o}^o, \boldsymbol{p}_{n_h}^h, \boldsymbol{n}_{n_h}^h] \in \mathbb{R}^9$. The network takes this feature point cloud $\mathcal{P}_{\mathcal{HOI}} \in \mathbb{R}^{|\mathcal{X}| \times (3+9)}$ as input. Subsequently, a PointTransformer~\cite{wu2024ptv3} encoder serves as a sparse backbone to extract local geometric embeddings $\mathcal{U}^{\text{geo}} \in \mathbb{R}^{|\mathcal{X}| \times C_{\text{geo}}}$ from this point cloud. To handle the variable number of contacts, we introduce a \ac{camp} mechanism. It computes max, mean, and standard-deviation statistics, denoted as $\boldsymbol{u}_{\max}$, $\boldsymbol{u}_{\mathrm{mean}}$, and $\boldsymbol{u}_{\mathrm{std}}$, and dynamically fuses them through a learnable weight vector:
\begin{equation}
    \label{eq:base_contact_aware_pooling_weight}
    \begin{aligned}
    \boldsymbol{w}
    =
    \mathrm{Softmax}
    \left(
    \psi
    \left(
    [
    \boldsymbol{u}_{\max},
    \boldsymbol{u}_{\mathrm{mean}},
    \boldsymbol{u}_{\mathrm{std}},
    \phi(|\mathcal{X}|)
    ]
    \right)
    \right),
    \end{aligned}
\end{equation}
where $\phi(\cdot)$ and $\psi(\cdot)$ are \acp{mlp} that encode the contact cardinality and project the fused features. The global contact descriptor is
\begin{equation}
    \label{eq:base_contact_descriptor}
    \begin{aligned}
    \boldsymbol{z}
    =
    w_1\boldsymbol{u}_{\max}
    +
    w_2\boldsymbol{u}_{\mathrm{mean}}
    +
    w_3\boldsymbol{u}_{\mathrm{std}}.
    \end{aligned}
\end{equation}
Finally, an \ac{mlp} maps $\boldsymbol{z}$ to a scalar stability score:
\begin{equation}
    \label{eq:base_decision_score_instantiation}
    \begin{aligned}
    \beta
    =
    \sigma
    \left(
    \mathrm{MLP}(\boldsymbol{z})
    \right),
    \end{aligned}
\end{equation}
where $\sigma(\cdot)$ denotes the sigmoid function. To train the grasp decision model, we use the \ac{bce} loss, with the optimization objective defined as:
\begin{equation}
    \mathcal{L}_{\text{BCE}} = - \left[ y_i \log(\beta_i) + (1 - y_i) \log(1 - \beta_i) \right],
\end{equation}
where $y_i \in \{0, 1\}$ represents the ground truth label, and $\beta_i$ denotes the predicted probability for the $i$-th sample.

\section{Foundation Priors for Adaptive Grasping}
\label{sec:foundation_prior_instantiation}

This section instantiates the transformation function $f_{\tau}(\cdot)$ with composable foundation priors. We consider three tasks with increasing complexity: grasping in cluttered scenes, functional grasping in cluttered scenes, and functional grasping in dynamic cluttered scenes. For these tasks, the transformation function is dynamically configured as
\begin{equation}
    \label{eq:prior_tau_composition}
    \begin{aligned}
    f_{\tau}
    =
    \begin{cases}
    \mathcal{M}_{\mathrm{s}}, & \text{cluttered grasping}, \\
    \mathcal{M}_{\mathrm{c}}\circ\mathcal{M}_{\mathrm{s}}, & \text{functional grasping}, \\
    \mathcal{M}_{\mathrm{t}}\circ\mathcal{M}_{\mathrm{c}}\circ\mathcal{M}_{\mathrm{s}}, & \text{dynamic grasping}.
    \end{cases}
    \end{aligned}
\end{equation}
Here, $\mathcal{M}_{\mathrm{s}}$, $\mathcal{M}_{\mathrm{c}}$, and $\mathcal{M}_{\mathrm{t}}$ denote the spatial-prior (\cref{subsec:spatial_prior_cluttered_grasping}), cognitive-prior (\cref{subsec:cognitive_prior_functional_grasping}), and temporal-prior (\cref{subsec:temporal_prior_dynamic_grasping}) modules, respectively, all of which can be flexibly instantiated by diverse composable foundation priors.

\begin{figure*}[t!]
    \centering
    \includegraphics[width=\linewidth]{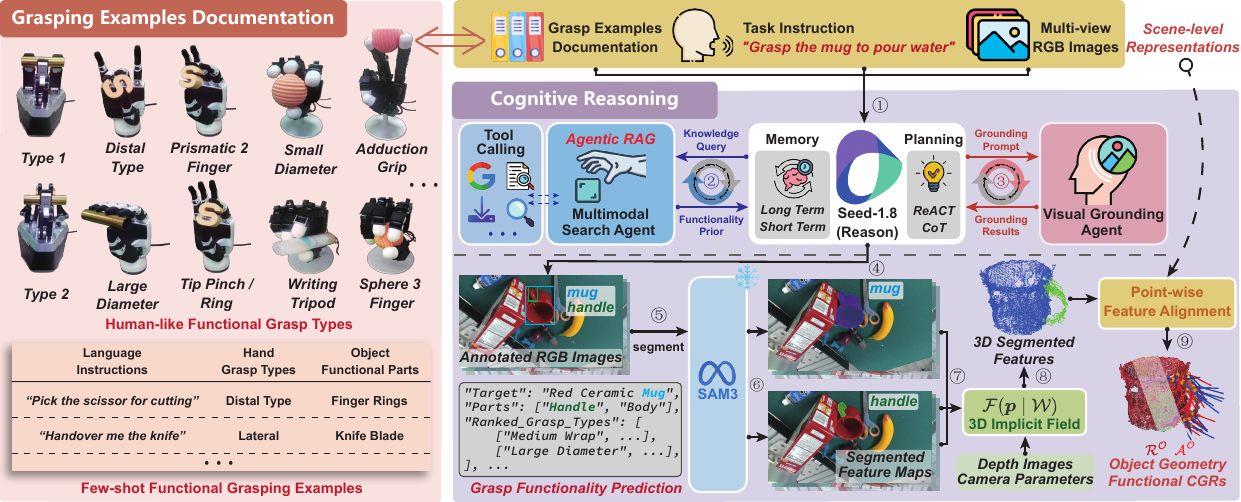}
    \caption{\textbf{Cognitive prior module.} Given scene observations and a language instruction, the cognitive prior module uses an \ac{llm} with \ac{rag}~\cite{lewis2020retrieval} and \ac{cot}~\cite{wei2022chain} mechanism to infer the target object, functional part, and hand-compatible grasp types. It then employs SAM3~\cite{carion2025sam3} and 3D lifting to align these inferred results with scene representations, producing the target object geometry $\mathcal{O}$, object-level semantics $\mathcal{R}^{\mathcal{O}}$, and functional \acp{cgr} $\mathcal{A}^{\mathcal{O}}$. These outputs, together with $\mathcal{T}_{\mathcal{H}}$, provide the task-conditioned interface for the base grasp policy to perform language-guided functional grasping.}
    \label{fig:cognitive_reasoning}
\end{figure*}

\subsection{Spatial Prior for Cluttered Grasping}
\label{subsec:spatial_prior_cluttered_grasping}
For grasping in cluttered scenes, where multiple objects are densely stacked in an unorganized manner, the task context consists of the multi-view scene observations $\mathcal{S}$:

\begin{equation}
    \label{eq:prior_spatial_condition}
    \begin{aligned}
    \mathcal{C}_{\mathrm{s}}
    =
    \{\mathcal{S}\},
    \;
    \mathcal{S}
    =
    \{I_m,D_m\}_{m=1}^{N_v},
    \end{aligned}
\end{equation}
where $I_m$ and $D_m$ denote the RGB image and depth map from the $m$-th camera view. Given the scene observations, the spatial prior module $\mathcal{M}_{\mathrm{s}}(\cdot)$ extracts scene semantic representations $\mathcal{R}^{\mathcal{S}}$ and scene-level \ac{cgr} candidates $\mathcal{A}^{\mathcal{S}}$:
\begin{equation}
    \label{eq:prior_spatial_mapping}
    \begin{aligned}
    \mathcal{M}_{\mathrm{s}}:
    \mathcal{S}
    \mapsto
    \left(
    \mathcal{R}^{\mathcal{S}},
    \mathcal{A}^{\mathcal{S}}
    \right),
    \end{aligned}
\end{equation}
where $\mathcal{R}^{\mathcal{S}}=\{\boldsymbol{r}_{n_s}\}_{n_s=1}^{|\mathcal{R}^{\mathcal{S}}|}$ and $\mathcal{A}^{\mathcal{S}}=\{\boldsymbol{a}_{j_s}\}_{j_s=1}^{|\mathcal{A}^{\mathcal{S}}|}$. These outputs provide the geometry and feasible contact primitives required for grasping in clutter. The pipeline is depicted in \cref{fig:spatial_understanding}.

\subsubsection{Scene Semantics Construction}
\label{subsubsec:prior_scene_semantic_representations}
To construct point-wise scene semantic representations, each RGB image is fed into the DINOv3~\cite{simeoni2025dinov3} encoder to extract pixel-level semantic feature maps, denoted as $\mathcal{W}^{\mathrm{sem}}=\{\boldsymbol{w}^{\mathrm{sem}}_m\}_{m=1}^{N_v}$. Here, $\boldsymbol{w}^{\mathrm{sem}}_m\in\mathbb{R}^{H\times W\times C_{\mathrm{sem}}}$ denotes the semantic feature map of the $m$-th view, where $H$ and $W$ are the image height and width, and $C_{\mathrm{sem}}$ is the semantic feature dimension.

We then lift these 2D semantic features into 3D space by using an implicit field function $\mathcal{F}(\cdot\mid\mathcal{W})$~\cite{wang2023d3ield}. This function maps an arbitrary 3D point to a feature vector by querying dense multi-view image features:
\begin{equation}
    \label{eq:prior_d3field_function}
    \boldsymbol{u}
    =
    \mathcal{F}(\boldsymbol{p}\mid\mathcal{W}),
\end{equation}
where $\boldsymbol{p}\in\mathbb{R}^{3}$ denotes a spatial point and $\boldsymbol{u}\in\mathbb{R}^{C}$ is its corresponding feature vector. For each point $\boldsymbol{p}_{n_s}^{s}\in\mathbb{R}^{3}$ in the scene point cloud $\mathcal{P}_{\mathcal{S}}\in\mathbb{R}^{N_s\times3}$, we query the semantic implicit field constructed from $\mathcal{W}^{\mathrm{sem}}$ to obtain its semantic descriptor:
\begin{equation}
    \label{eq:prior_scene_semantic_representation}
    \boldsymbol{r}_{n_s}
    =
    \mathcal{F}
    \left(
    \boldsymbol{p}_{n_s}^{s}
    \mid
    \mathcal{W}^{\mathrm{sem}}
    \right).
\end{equation}
Aggregating the semantic descriptors of all scene points yields the scene semantic representations $\mathcal{R}^{\mathcal{S}}$.

\subsubsection{Scene-level \acp{cgr} Prediction}
\label{subsubsec:prior_scene_cgr_prediction}
To predict scene-level \acp{cgr} from the multi-view scene observations, we adopt the PyTorch implementation of Contact-GraspNet\footnote{\url{https://github.com/SeungBack/gc6d-contact-graspnet}}. We train this network on the GraspClutter6D~\cite{back2025graspclutter6d} dataset, retaining the original hyperparameter settings. Given the scene point cloud $\mathcal{P}_{\mathcal{S}}$ and semantic representations $\mathcal{R}^{\mathcal{S}}$, the network produces dense per-point \acp{cgr} $\{\boldsymbol{a}_{n_s}\}$. During inference, we reconstruct the scene-level \ac{cgr} set by filtering these dense predictions with the contact probability threshold $\delta_c$:
\begin{equation}
    \label{eq:prior_scene_cgr_filtered}
    \begin{aligned}
    \mathcal{A}^{\mathcal{S}}
    =
    \left\{
    \boldsymbol{a}_{n_s}
    \mid
    p_{n_s}>\delta_c
    \right\}
    =
    \{\boldsymbol{a}_{j_s}\}.
    \end{aligned}
\end{equation}
Together with the scene geometry and all available grasp types, these \acp{cgr} construct the task-conditioned interface for clutter-aware grasping. This interface further serves as the input to the cognitive prior, where the most suitable grasp type and target region are selected according to the task instruction.

\subsection{Cognitive Prior for Functional Grasping}
\label{subsec:cognitive_prior_functional_grasping}
For functional grasping in cluttered scenes, the task context further includes a natural language instruction $\ell$:
\begin{equation}
    \label{eq:prior_cognitive_condition}
    \begin{aligned}
    \mathcal{C}_{\mathrm{c}}
    =
    \{\mathcal{S},\ell\}.
    \end{aligned}
\end{equation}
Given the multi-view scene observations and language instruction, the cognitive prior module $\mathcal{M}_{\mathrm{c}}(\cdot)$ infers the target object, functional part, and appropriate grasp types, and then aligns these constraints with 3D scene representations:
\begin{equation}
    \label{eq:prior_cognitive_mapping}
    \begin{aligned}
    \mathcal{M}_{\mathrm{c}}:
    \left(
    \mathcal{S},
    \ell,
    \mathcal{R}^{\mathcal{S}},
    \mathcal{A}^{\mathcal{S}}
    \right)
    \mapsto
    \left(
    \mathcal{O},
    \mathcal{T}_{\mathcal{H}},
    \mathcal{R}^{\mathcal{O}},
    \mathcal{A}^{\mathcal{O}}
    \right),
    \end{aligned}
\end{equation}
where $\mathcal{R}^{\mathcal{O}}=\{\boldsymbol{r}_{n_o}\}_{n_o=1}^{|\mathcal{R}^{\mathcal{O}}|}$ represents the set of object semantic representations extracted from the scene representations $\mathcal{R}^{\mathcal{S}}$, while $\mathcal{A}^{\mathcal{O}}=\{\boldsymbol{a}_{j_o}\}_{j_o=1}^{|\mathcal{A}^{\mathcal{O}}|}$ denotes the functional \acp{cgr} aligned with the grounded target object and the functional part.

\subsubsection{Grasp Functionality Inference}
\label{subsubsec:prior_grasp_functionality_inference}
To understand human language instructions, we instantiate a parsing function $\Psi(\cdot)$ with an \ac{llm} to infer the structured grasp specification:
\begin{equation}
    \label{eq:prior_language_parsing}
    \begin{aligned}
    \left(
    \ell_{\mathrm{obj}},
    \ell_{\mathrm{part}},
    \mathcal{T}_{\mathcal{H}}
    \right)
    =
    \Psi
    \left(
    \mathcal{S},
    \ell
    \right),
    \end{aligned}
\end{equation}
where $\ell_{\mathrm{obj}}$ and $\ell_{\mathrm{part}}$ denote the language descriptors for the target object and its functional part. Direct inference of these physically grounded attributes is non-trivial due to the inherent lack of grasp reasoning in standard \acp{llm}. To bridge this gap, we employ Seed-1.8\footnote{\url{https://github.com/ByteDance-Seed/Seed-1.8}} as the core reasoning agent, augmented with \ac{rag}~\cite{lewis2020retrieval} and \ac{cot}~\cite{wei2022chain} mechanisms.

As shown in \cref{fig:cognitive_reasoning}, we construct a grasp taxonomy database that encodes detailed attributes for predefined grasp types, including functional intent, active finger configurations, contact constraints, and use-case examples. By retrieving relevant knowledge from this database, we constrain the inference space and mitigate hallucinations in \acp{llm}. Furthermore, we use \ac{cot} prompting to decompose the inference into task intent identification, functional part selection, and grasp requirement analysis, ultimately outputting $\ell_{\mathrm{obj}}$, $\ell_{\mathrm{part}}$, and $\mathcal{T}_{\mathcal{H}}$.

\subsubsection{Object Semantic Representation Refinement}
\label{subsubsec:prior_object_semantic_refinement}
Then, we ground the parsed object description $\ell_{\mathrm{obj}}$ in the scene semantic representations. To localize the target object, we perform visual grounding conditioned on the parsed object description, such as \textit{``the mug''}. Specifically, we use Seed-1.8 to generate bounding-box annotations on multi-view RGB images. These annotations are then used as prompts for SAM3~\cite{carion2025sam3} to extract multi-view instance masks, denoted as $\mathcal{W}^{\mathrm{inst}}=\{\boldsymbol{w}^{\mathrm{inst}}_m\}_{m=1}^{N_v}$.

We further lift the 2D instance masks into 3D space using the implicit field function defined in \cref{eq:prior_d3field_function}, yielding a point-wise instance probability $\boldsymbol{u}^{\mathrm{inst}}_{n_s}\in[0,1]$ for each scene point. To obtain object semantic representations, we filter the scene-level representations and retain only those associated with the grounded target instance:
\begin{equation}
    \label{eq:prior_object_semantic_refine}
    \begin{aligned}
    \mathcal{R}^{\mathcal{O}}
    \triangleq
    \left\{
    \boldsymbol{r}_{n_s}\in\mathcal{R}^{\mathcal{S}}
    \mid
    \boldsymbol{u}^{\mathrm{inst}}_{n_s}>\delta_{\mathrm{inst}}
    \right\}
    =
    \{\boldsymbol{r}_{n_o}\}.
    \end{aligned}
\end{equation}
The corresponding target object geometry $\mathcal{O}$ is extracted from the scene point cloud using the same grounded instance mask.

\subsubsection{Functional \acp{cgr} Alignment}
\label{subsubsec:prior_functional_cgr_alignment}
We further use the parsed part description $\ell_{\mathrm{part}}$, such as \textit{``the handle''}, to restrict grasp generation to the corresponding functional regions. Analogous to the instance grounding, we derive a point-wise functional probability $\boldsymbol{u}^{\mathrm{fun}}_{n_s}\in[0,1]$ by identifying functional regions through visual grounding and 3D lifting. To extract task-relevant contact primitives, we filter the scene-level \ac{cgr} set $\mathcal{A}^{\mathcal{S}}$ by enforcing both instance-level and functional-level consistency:
\begin{equation}
    \label{eq:prior_functional_cgr_alignment}
    \begin{aligned}
    \mathcal{A}^{\mathcal{O}}
    \triangleq
    \left\{
    \boldsymbol{a}_{n_s}\in\mathcal{A}^{\mathcal{S}}
    \mid
    \boldsymbol{u}^{\mathrm{inst}}_{n_s}>\delta_{\mathrm{inst}}
    \land
    \boldsymbol{u}^{\mathrm{fun}}_{n_s}>\delta_{\mathrm{fun}}
    \right\}= \left\{ \boldsymbol{a}_{j_o} \right\},
    \end{aligned}
\end{equation}
This set provides functionally relevant and spatially feasible \ac{cgr} candidates for grasp synthesis. Together with the object geometry $\mathcal{O}$ and the inferred grasp types $\mathcal{T}_{\mathcal{H}}$, it forms the task-conditioned interface for language-guided functional grasping.

\begin{figure}[t!]
    \centering
    \includegraphics[width=\linewidth]{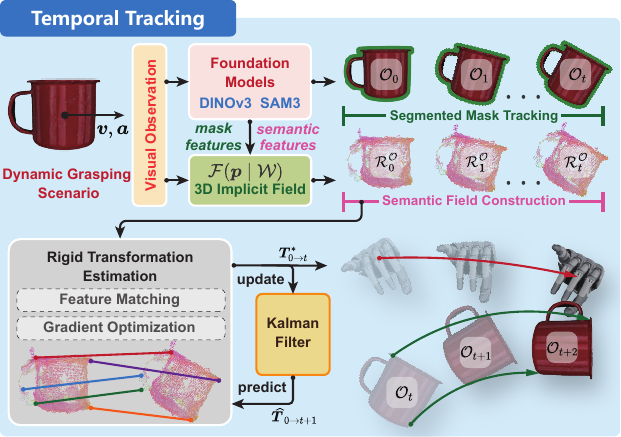}
    \caption{\textbf{Temporal prior module.} This module integrates the temporal mask-tracking mechanism of SAM3~\cite{carion2025sam3} with the cross-frame consistency of DINOv3~\cite{simeoni2025dinov3} features to estimate the target's rigid motion. The estimated motion is then used to update the structured interface, enabling the base grasp policy to perform temporally consistent dynamic grasping.}
    \label{fig:temporal_tracking}
\end{figure}

\subsection{Temporal Prior for Dynamic Grasping}
\label{subsec:temporal_prior_dynamic_grasping}
For functional grasping in dynamic cluttered scenes, the task context comprises a continuous observation sequence and a language instruction:
\begin{equation}
    \label{eq:prior_temporal_condition}
    \begin{aligned}
    \mathcal{C}_{\mathrm{t}}
    =
    \{\mathcal{S}_{0:t},\ell\}.
    \end{aligned}
\end{equation}
At the initial time step $t=0$, the spatial and cognitive priors construct the initial grasp interface:
\begin{equation}
    \label{eq:prior_initial_interface}
    \begin{aligned}
    \mathcal{I}_{0}
    =
    \left(
    \mathcal{O}_{0},
    \mathcal{A}^{\mathcal{O}}_{0},
    \mathcal{T}_{\mathcal{H}}
    \right).
    \end{aligned}
\end{equation}
The temporal prior module $\mathcal{M}_{\mathrm{t}}(\cdot)$ updates this interface across continuous observations to maintain temporally consistent grasp tracking:
\begin{equation}
    \label{eq:prior_temporal_mapping}
    \begin{aligned}
    \mathcal{M}_{\mathrm{t}}:
    \left(
    \mathcal{I}_{0},
    \mathcal{S}_{1:t}
    \right)
    \mapsto
    \mathcal{I}_{t}
    =
    \left(
    \mathcal{O}_{t},
    \mathcal{A}_{t},
    \mathcal{T}_{\mathcal{H}}
    \right).
    \end{aligned}
\end{equation}
Under the rigid-body assumption, if a grasp is stable at the initial time step, it remains stable at subsequent time steps provided that the hand strictly tracks the object's rigid motion. Therefore, we instantiate $\mathcal{M}_{\mathrm{t}}(\cdot)$ to estimate the real-time $\mathrm{SE}(3)$ transformation of the target object. The estimated transformation is subsequently applied to update the grasp interface, providing temporally consistent inputs for the base grasp policy. The pipeline is depicted in \cref{fig:temporal_tracking}.

\subsubsection{Object Tracking Consistency}
\label{subsubsec:prior_object_tracking_consistency}
To maintain object tracking consistency, we leverage the mask-based tracking mechanism of SAM3~\cite{carion2025sam3}. Instead of performing independent visual grounding at each time step, we initialize the target instance mask $\mathcal{W}^{\mathrm{inst}}_0$ at the initial time step and propagate it over time. This mask-based tracking strategy effectively mitigates sequential segmentation instability and semantic drift, ensuring that the tracked instance mask $\mathcal{W}^{\mathrm{inst}}_t$ remains strictly consistent with the target object throughout the grasping process. Subsequently, $\mathcal{O}_t$ is extracted from the real-time scene observation and $\mathcal{W}^{\mathrm{inst}}_t$.

\subsubsection{Object Pose Estimation}
\label{subsubsec:prior_object_pose_estimation}
To estimate the target's rigid transformation matrix $\boldsymbol{T}^{\mathcal{O}}_{0\to t} \in\mathrm{SE}(3) $, we construct a semantic consistency optimization problem~\cite{wang2023d3ield}. Specifically, we denote the object's point cloud at the initial time step as $\mathcal{P}_{\mathcal{O},0} = \{\boldsymbol{p}_{n_o,0}^o\}_{n_o=1}^{N_o}$, and the corresponding set of semantic descriptors as $\mathcal{R}^{\mathcal{O}}_0 = \{\boldsymbol{r}_{n_o,0}^o\}_{{n_o}=1}^{N_o}$. At time step $t$, we compute the optimal rotation $\boldsymbol{R}_{0 \to t}^*$ and translation $\boldsymbol{t}_{0 \to t}^*$ that minimize the discrepancy between the initial semantic descriptors and the features queried from the current scene's implicit field $\mathcal{F}(\cdot \mid \mathcal{W}^{\text{sem}}_t)$. The optimization objective is defined as:
\begin{equation}
    \label{eq:prior_d3field_optimization_objective}
    \begin{aligned}
    &
    \left(
    \boldsymbol{R}_{0\to t}^{*},
    \boldsymbol{t}_{0\to t}^{*}
    \right)
    =
    \underset{
    \boldsymbol{R}_{0\to t},
    \boldsymbol{t}_{0\to t}
    }{\arg\min}
    \sum_{n_o=1}^{N_o}
    \left\|
    \mathcal{F}
    \left(
    \hat{\boldsymbol{p}}_{n_o,t}^{o}
    \mid
    \mathcal{W}^{\mathrm{sem}}_{t}
    \right)
    -
    \boldsymbol{r}_{n_o,0}^{o}
    \right\|_2^2, \\
    &
    \hat{\boldsymbol{p}}_{n_o,t}^{o}
    =
    \boldsymbol{R}_{0\to t}
    \boldsymbol{p}_{n_o,0}^{o}
    +
    \boldsymbol{t}_{0\to t}.
    \end{aligned}
\end{equation}
Since the implicit field function $\mathcal{F}(\cdot\mid\mathcal{W})$ is differentiable, this optimization can be solved using gradient-based methods. In practice, we further incorporate rigid-body constraints and distance regularization to improve robustness and stability.

Consequently, given the estimated target object's rotation and translation $(\boldsymbol{R}_{0 \to t}^*, \boldsymbol{t}_{0 \to t}^*)$, the real-time contact interface can be directly computed from the initial contact reference:
\begin{equation}
    \label{eq:prior_temporal_interface_update}
    \begin{aligned}
    \mathcal{A}_{t}
    =
    \left\{
        \left[
                \begin{array}{cccc} 
                    & \boldsymbol{R}_{0 \to t}^* & & \boldsymbol{t}_{0 \to t}^* \\ 
                    0 & 0 & 0 & 1 
                \end{array}
            \right] 
    \boldsymbol{a}
    \mid
    \boldsymbol{a} \in \mathcal{A}_{0}
    \right\}.
    \end{aligned}
\end{equation}

In practice, we observe that the pose-estimation error remains within an admissible tolerance. Therefore, to maximize system efficiency, we bypass the base grasp policy and directly update the initial grasp pose $\boldsymbol{g}_0 = (\boldsymbol{T}_0, \boldsymbol{q}_0)$ via the rigid transformation:
\begin{equation}
    \label{eq:prior_dynamic_grasp_update}
    \begin{aligned}
        \boldsymbol{g}_t = \left( \left[
            \begin{array}{cccc} 
                & \boldsymbol{R}_{0 \to t}^* & & \boldsymbol{t}_{0 \to t}^* \\ 
                0 & 0 & 0 & 1 
            \end{array}
        \right] \boldsymbol{T}_0, \boldsymbol{q}_0 \right).
    \end{aligned}
\end{equation}

\section{Simulated Experiments}
\label{sec:sim_exp}
In this section, we systematically validate the key components of \model in simulation. We first introduce the data collection and training setup (\cref{subsec:sim_exp_base_policy_training}), and then evaluate the base policy for generalizable grasp synthesis (\cref{subsec:sim_exp_grasp_decision_model}), the spatial prior for cluttered grasping (\cref{subsec:sim_exp_spatial_prior}), the cognitive prior for functional grasping (\cref{subsec:sim_exp_cognitive_prior}), and the temporal prior for dynamic grasping (\cref{subsec:sim_exp_temporal_prior}).

\subsection{Data Collection and Training Implementation}
\label{subsec:sim_exp_base_policy_training}

\begin{figure}[t!]
    \centering
    \includegraphics[width=\linewidth]{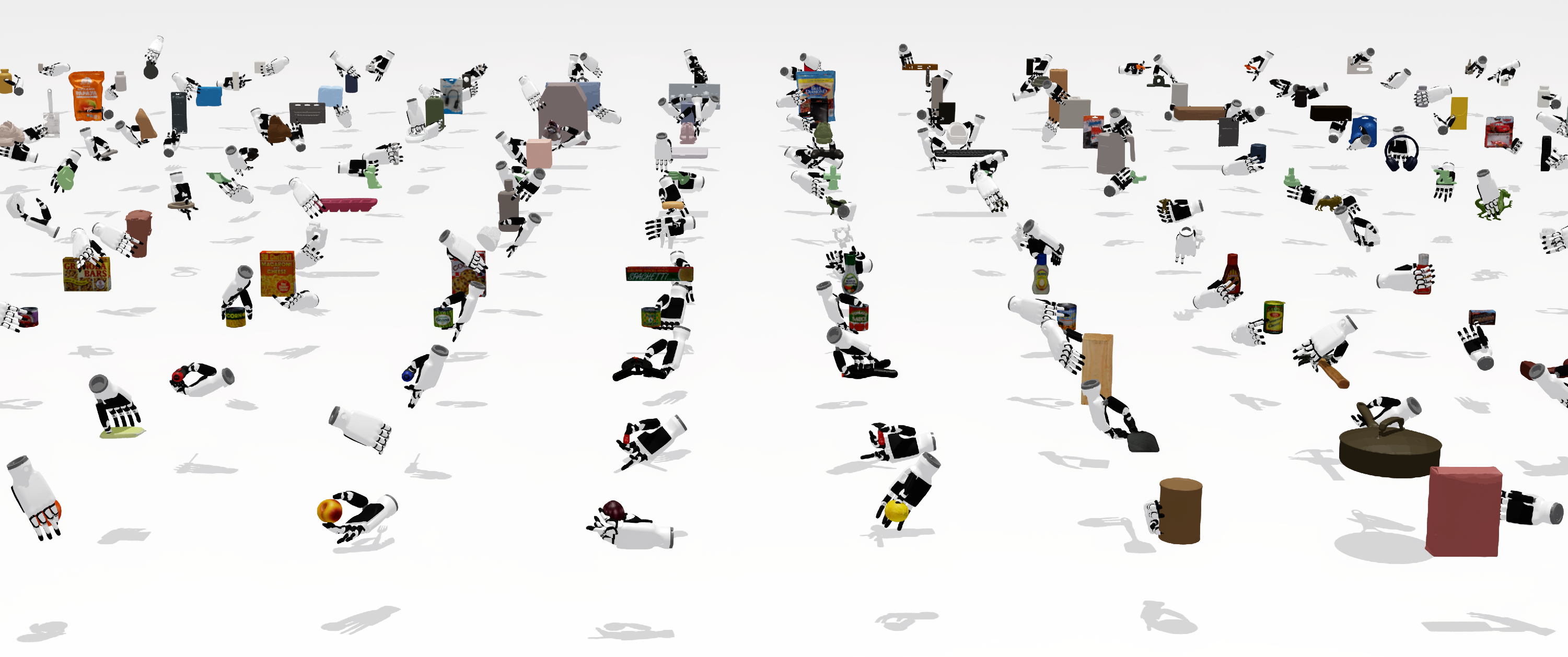}
    \caption{\textbf{Data collection in simulation.} To train and evaluate the grasp decision model, we collect synthetic data via trial and error in Isaac Sim. The dataset comprises 4.4 million annotated grasping trials across three diverse robotic hands.}
    \label{fig:sim_data_collection}
\end{figure}

\subsubsection{Data Collection and Annotation}
To train the grasp decision model $\mathcal{D}(\cdot)$, we collect data via trial and error on single-object grasping in a physics-based simulator (see \cref{fig:sim_data_collection}). All object models are sourced from the training sets of the GraspNet-1Billion~\cite{fang2020graspnet} and GraspClutter6D~\cite{back2025graspclutter6d} benchmarks. Our data generation pipeline consists of two stages.

In the first stage, we densely sample \ac{cgr} candidates on the object surface. Candidates satisfying basic force-closure criteria are then mapped into executable hand poses via $\mathcal{M}_{\mathcal{H}}(\cdot)$ defined in \cref{subsec:base_policy_hand_mapping}. In the second stage, we execute these proposals in Isaac Sim to evaluate grasp stability, with the friction coefficient set to 0.5. After the hand closes to establish contact, we apply a $\qty{0.5}{\newton}$ external perturbation force to the object along six axis-aligned directions. A grasp is labeled as successful ($y=1$) only if the object's translational displacement remains under $\qty{0.05}{\meter}$ and rotational deviation under $\ang{30}$; otherwise, it is labeled as a failure ($y=0$). We apply this pipeline to three diverse robotic hands (DH3, Allegro, and Inspire). For each hand, we collect 1,000 grasping trials for every combination of object and grasp type, balanced between positive and negative examples. With 200 training objects and the grasp types available for each hand (4 for DH3, 10 for Allegro, and 8 for Inspire), this process yields 4.4 million annotated grasping trials in total. We partition the dataset into training and validation sets at a 10:1 ratio.

\subsubsection{Training Implementation}
We implement the grasp decision model in PyTorch on Ubuntu 20.04, and train it on a server with dual Intel Xeon 6330 28-core CPUs, four NVIDIA GeForce RTX 4090 GPUs, and 256 GB of RAM. We use the Adam optimizer with a learning rate of $1\times10^{-4}$ and a weight decay of $0.05$. The model is trained for 30 epochs with a batch size of 512, using the binary cross-entropy loss and a cosine annealing learning rate scheduler.

Throughout training and inference, the feasible grasp types for each hand follow the setup in AnyDexGrasp~\cite{fang2025anydexgrasp}. To enhance robustness against real-world observation noise, we inject Gaussian perturbations ($\sigma = \qty{0.005}{\meter}$ for coordinates, $\sigma = 0.02$ for normals) during model training. During inference, a grasp is considered stable if the predicted score $\beta > 0.5$.

\subsection{Evaluation of the Base Policy for General Grasp Synthesis}
\label{subsec:sim_exp_grasp_decision_model}

The base policy consists of two stages: hand-specific grasp mapping and hand-agnostic grasp evaluation. The first stage is instantiated as an explicit function and supports deployment across end-effectors with two to five fingers, as demonstrated in subsequent experiments. Therefore, we focus on evaluating the proposed \ac{hoi} representation and grasp decision model on learning efficiency and cross-hand grasp evaluation.

\begin{figure}[t!]
    \centering
    \begin{subfigure}{0.49\columnwidth}
        \centering
        \includegraphics[width=\linewidth]{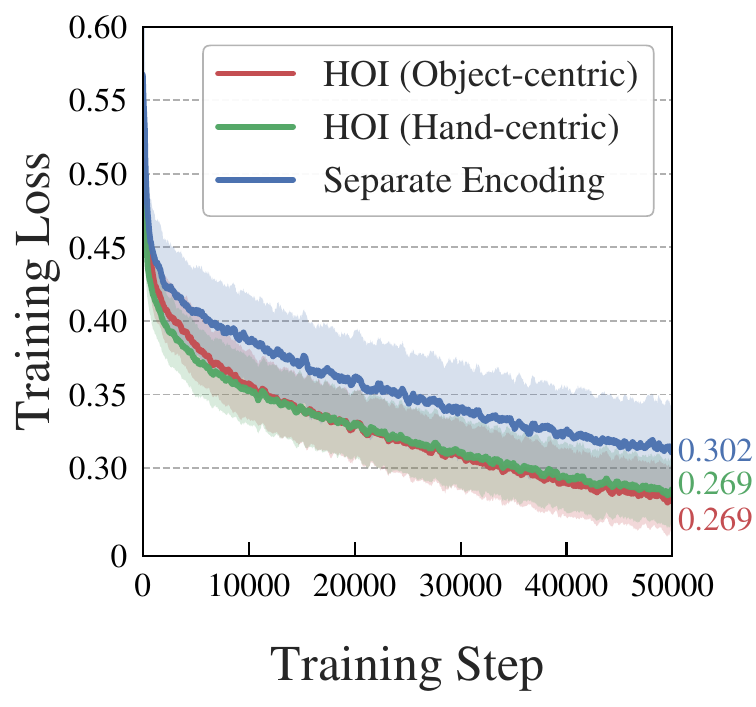}
        \caption{Training convergence.}
        \label{fig:sim_gd_training_convergence}
    \end{subfigure}
    \hfill
    \begin{subfigure}{0.49\columnwidth}
        \centering
        \includegraphics[width=\linewidth]{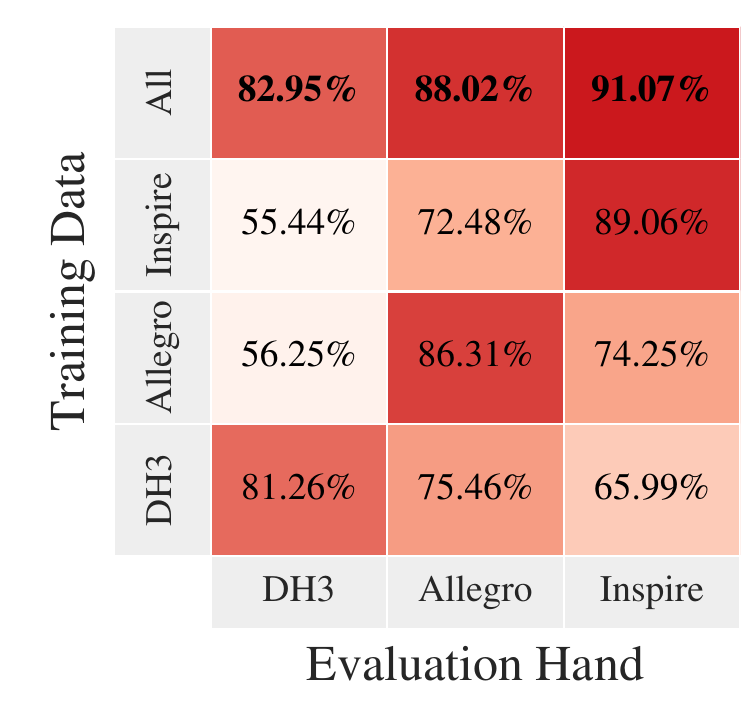}
        \caption{Cross-hand accuracy.}
        \label{fig:sim_gd_cross_hand_accuracy}
    \end{subfigure}
    \caption{\textbf{Comparative experiments of the grasp decision model.} (a) Training loss of the grasp decision model using different representations. (b) Grasp quality prediction accuracy of the grasp decision model trained on different hand data.}
    \label{fig:grasp_decision_model_comparison}
\end{figure}

\begin{figure}[t!]
    \centering
    \includegraphics[width=\linewidth]{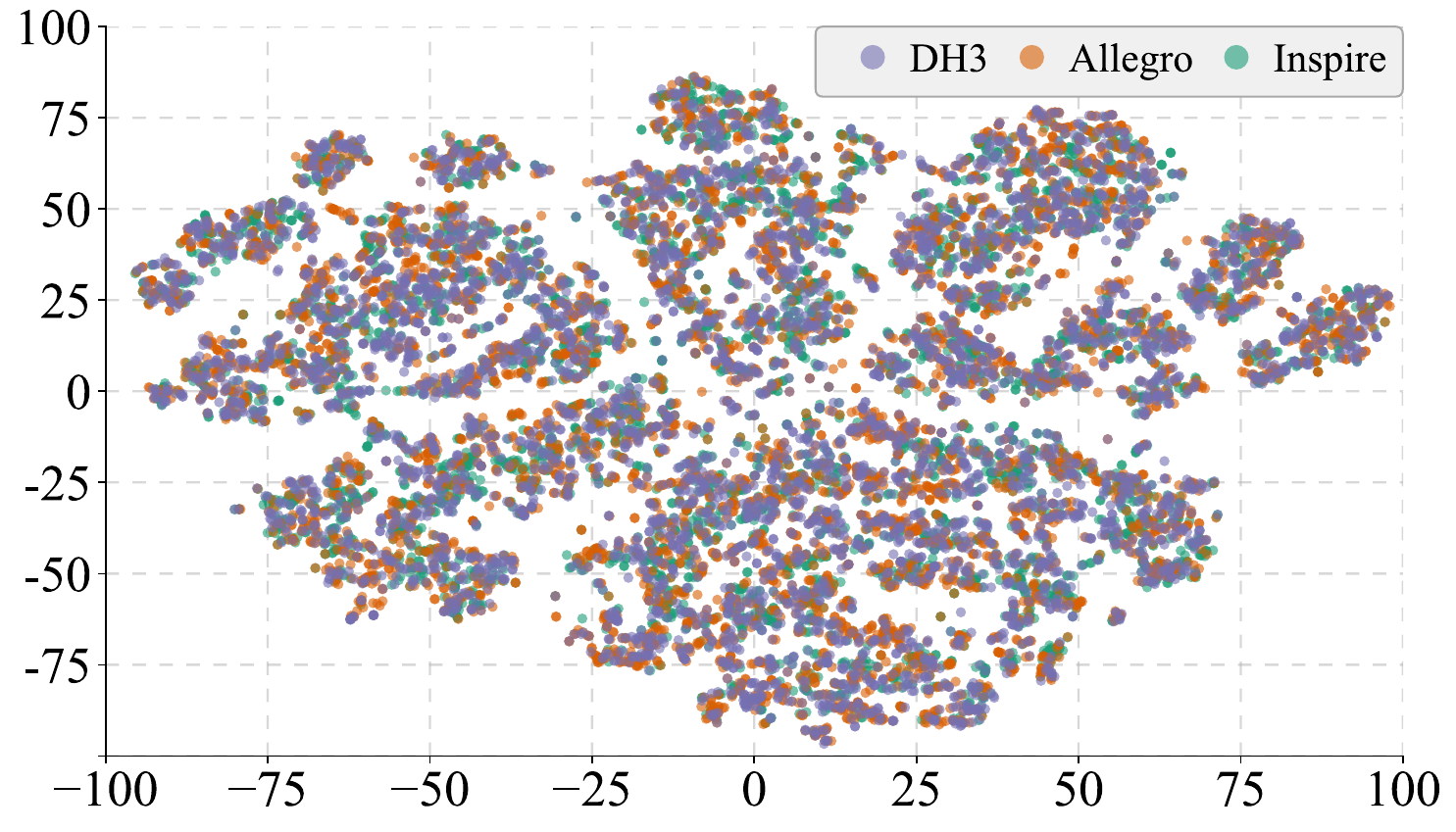}
    \caption{\textbf{t-SNE visualization of \ac{hoi} representations.} This figure visualizes the distributions of stable \ac{hoi} representations across three different robotic hands (DH3, Allegro, and Inspire).}
    \label{fig:hoi_tsne_vis}
\end{figure}

\subsubsection{Learning Efficiency}
To evaluate the impact of different input representations on the learning efficiency of the decision model, we train the model on the Allegro hand using the synthetic grasping data constructed in \cref{subsec:sim_exp_base_policy_training}. We compare three input paradigms: object-centric \ac{hoi}, hand-centric \ac{hoi}, and an explicitly separated hand-object point cloud representation. The former two encode the \ac{hoi} relationship from object-centric and hand-centric perspectives, respectively, while the latter independently encodes the contact point clouds of the hand and object and then concatenates their features. All compared variants share the same network architecture, and convergence behavior is evaluated through the training loss curves.

As illustrated in \cref{fig:sim_gd_training_convergence}, compared with the separated representation, the \ac{hoi} representations lead to faster network convergence and lower final training losses. This improvement is mainly attributed to the ability of the \ac{hoi} representation to abstract the complex force-closure condition into a compact feature point cloud, thereby substantially reducing the learning difficulty for the network.

\subsubsection{Cross-hand Generalization}
To evaluate whether the \ac{hoi} representation enables cross-hand generalization of the decision model, we train the decision model using single-hand data and the combined data from all three hands, respectively. The grasp quality prediction accuracy is then evaluated across the three robotic hands.

As shown in \cref{fig:sim_gd_cross_hand_accuracy}, the model jointly trained on data from all three robotic hands achieves accuracies of 82.95\%, 88.02\%, and 91.07\% on the DH3, Allegro, and Inspire, respectively, outperforming the single-hand training configurations overall. This result demonstrates that the \ac{hoi} representation successfully facilitates shared stability judgment across different robotic hands, thereby enhancing cross-hand generalization capabilities.

Furthermore, we provide a t-SNE visualization of stable \ac{hoi} representations across the three robotic hands (see \cref{fig:hoi_tsne_vis}). The feature distributions exhibit substantial overlap in the shared embedding space, with no obvious clustering induced by kinematic or structural differences. This observation is consistent with the hand-agnostic property of the \ac{hoi} representation and provides qualitative evidence for its suitability for joint multi-hand training and cross-hand generalization.

\subsection{Evaluation of Spatial Prior for Clutter-Aware Grasping}
\label{subsec:sim_exp_spatial_prior}
This experiment evaluates the effectiveness of the spatial prior in transforming multi-view observations into scene-level contact constraints for grasp synthesis, supporting spatially feasible grasp generation under different levels of scene clutter.

\subsubsection{Benchmark Setting}
We adopt the simulated benchmark proposed in DexGraspNet 2.0~\cite{zhang2024dexgraspnet}, which is built upon the GraspNet-1Billion dataset~\cite{fang2020graspnet}, and comprises 450 scenes and 88 diverse objects. To evaluate the model performance under different clutter levels, the test scenes are categorized into three difficulty splits based on object density:
\begin{itemize}
    \item \textit{Dense} consists of 90 scenes from the test set in GraspNet-1Billion~\cite{fang2020graspnet}, with each scene containing 8-11 objects.
    \item \textit{Random} comprises 180 scenes by augmenting the dense split twice, where each scene retains 1-10 random objects.
    \item \textit{Loose} contains 180 scenes by augmenting the dense split twice, with 1-2 random objects remaining in the scene.
\end{itemize}
All experiments are conducted in Isaac Gym, where the friction coefficient is set to $\mu=0.2$.

\subsubsection{Evaluation Metrics}
We report the grasp \textit{success rate}, following the same evaluation protocol as DexGraspNet 2.0~\cite{zhang2024dexgraspnet}. Specifically, a grasp trial is considered a success if the highest-scoring grasp pose lifts the object off the table without any initial collision with the table or surrounding objects.

\subsubsection{Baseline Models}
\label{subsubsec:sim_exp_spatial_prior_baselines}
We compare \model against AnyDexGrasp~\cite{fang2025anydexgrasp} and its variant, AnyDexGrasp${}^*$. AnyDexGrasp is a two-stage dexterous grasping framework that first predicts a hand-agnostic grasp representation to generate hand-specific grasp candidates, and subsequently employs a hand-specific decision model to select executable grasps. Because it supports the same three robotic hands and adopts a grasp taxonomy consistent with ours, we construct AnyDexGrasp${}^*$ as a variant augmented with our grasp decision model.

\subsubsection{Experimental Details}
AnyDexGrasp, AnyDexGrasp${}^*$, and \model are evaluated in a zero-shot manner on three robotic hands spanning three to five fingers (DH3, Allegro, and Inspire), without any re-training. For AnyDexGrasp and AnyDexGrasp${}^*$, we use the officially released representation model to generate hand-agnostic grasp candidates. AnyDexGrasp scores these candidates with its original decision model, whereas AnyDexGrasp${}^*$ uses our grasp decision model for candidate ranking. During inference, the highest-scoring grasp pose is selected for execution.

\newcolumntype{Y}{>{\centering\arraybackslash}X}

\begin{table}[t!]
    \centering
    \small
    \caption{\textbf{Performance comparison of multi-finger grasp pose generation in clutter on the DexGraspNet 2.0~\cite{zhang2024dexgraspnet} benchmark.} We report the grasp success rates across three clutter difficulty levels (\ie, Dense, Random, and Loose), comparing various methods across different robotic hands. The best results are highlighted in bold and underlined.}
    \begin{tabularx}{\linewidth}{
        >{\hsize=2.0\hsize}X %
        >{\hsize=1.2\hsize}Y %
        >{\hsize=0.6\hsize}Y %
        >{\hsize=0.6\hsize}Y
        >{\hsize=0.6\hsize}Y
    } %
        \toprule
        \multirow{2}{*}{\makecell[l]{Methods}} & \multirow{2}{*}{\makecell[l]{Hands}}
        & \multicolumn{3}{c}{Success Rate (\%)} \\ 

        \cmidrule(lr){3-5}
        & & Dense & Random & Loose \\

        \midrule
        AnyDexGrasp~\cite{fang2025anydexgrasp} & DH3 & \underline{\textbf{91.2}} & 85.5 & 78.2 \\
        AnyDexGrasp${}^*$ & DH3 & 87.8 & 86.4 & 85.7 \\
        \model (ours) & DH3 & 90.3 & \underline{\textbf{89.5}} & \underline{\textbf{88.6}} \\

        \midrule
        AnyDexGrasp~\cite{fang2025anydexgrasp} & Allegro  & \underline{\textbf{90.5}} & 86.6 & 79.4 \\
        AnyDexGrasp${}^*$ & Allegro  & 88.2 & 87.2 & 83.8 \\
        \model (ours) & Allegro  & 89.5 & \underline{\textbf{88.2}} & \underline{\textbf{85.9}} \\

        \midrule
        AnyDexGrasp~\cite{fang2025anydexgrasp} & Inspire  & \underline{\textbf{92.7}} & 85.0 & 76.1 \\
        AnyDexGrasp${}^*$ & Inspire  & 86.1 & 85.4 & 83.9 \\
        \model (ours) & Inspire  & 88.1 & \underline{\textbf{88.5}} & \underline{\textbf{86.7}} \\
        
        \bottomrule
    \end{tabularx}
    \label{tab:sim_mf_grasp_clutter}
\end{table}

\subsubsection{Experimental Results}
As shown in \cref{tab:sim_mf_grasp_clutter}, \model exhibits strong zero-shot and cross-hand grasp synthesis in cluttered scenes. Compared with AnyDexGrasp, it achieves the highest success rates on the random and loose splits for all three hands, with average success-rate improvements of 3.0\% and 9.2\%. Although AnyDexGrasp performs slightly better on the dense split, \model shows more stable performance across different clutter levels. This primarily stems from compact \acp{cgr} and the human-inspired grasp taxonomy, which significantly reduce the grasp search space, thereby enabling better adaptation to scenes of varying complexity.

The comparison with AnyDexGrasp${}^*$ further validates our grasp decision model. Given the same grasp candidates as AnyDexGrasp, AnyDexGrasp${}^*$ improves the average success rate by 0.6\% on the random split and 6.6\% on the loose split. Moreover, \model consistently outperforms AnyDexGrasp${}^*$ across all three hands, indicating that both the spatial prior and the hand-agnostic grasp decision model jointly contribute to robust multi-finger grasp synthesis.

\subsection{Evaluation of Cognitive Prior for Functional Grasping}
\label{subsec:sim_exp_cognitive_prior}
This experiment evaluates the effectiveness of the cognitive prior in converting language instructions into functional grasp constraints for grasp synthesis. Because current simulation benchmarks fall short in jointly evaluating grasp functionality inference, target grounding, and physical grasp execution, we assess this prior module from two complementary experiments: grasp functionality inference on open-set instructions and language-guided target grasping in simulated clutter.

\subsubsection{Grasp Functionality Inference}

To evaluate grasp functionality inference, we construct an open-set instruction dataset based on instruction templates~\cite{tang2023graspgpt,tang2025foundationgrasp}. The dataset contains 125 image-instruction pairs spanning 57 manipulation tasks and 80 object categories, with human-annotated ground-truth functional grasp intents. We report functional part and grasp type accuracy as evaluation metrics.

We conduct an ablation study to evaluate the individual contributions of the \ac{cot} and \ac{rag} mechanisms in the cognitive reasoning module. \cref{tab:cognitive_ablation_exp} demonstrates that integrating \ac{rag} and \ac{cot} techniques substantially improves the model's grasp functionality inference accuracy. This efficient design eliminates the reliance on large-scale language-action paired data, thereby substantially reducing the overall learning cost.

\begin{table}[t!]
    \centering
    \small
    \caption{\textbf{Experimental results of grasp functionality inference on open-set instructions.} We report the accuracy of grasp functionality inference under different settings. The best results are highlighted in bold and underlined.}
    \begin{tabular*}{\columnwidth}{@{\extracolsep{\fill}}lcc}
        \toprule
        Method & \makecell{Functional Part\\Accuracy $\uparrow$} & \makecell{Grasp Type\\Accuracy $\uparrow$} \\
        \midrule
        Base LLM               & 0.82 & 0.63 \\
        Base LLM + \ac{cot}    & 0.91 & 0.66 \\
        Base LLM + \ac{rag}    & 0.85 & 0.78 \\
        \model (ours)          & \underline{\textbf{0.94}} & \underline{\textbf{0.87}} \\
        \bottomrule
    \end{tabular*}
    \label{tab:cognitive_ablation_exp}
\end{table}

\newcolumntype{Y}{>{\centering\arraybackslash}X}

\begin{table}[t!]
    \centering
    \small
    \caption{\textbf{Experimental results of language-guided target grasping for multi-finger hands on the GraspClutter6D~\cite{back2025graspclutter6d} and GraspNet-1Billion~\cite{fang2020graspnet} datasets.} We report the grasp success rates of three dexterous hands on two datasets. The best results are highlighted in bold and underlined.}
    \begin{tabularx}{\linewidth}{
        >{\hsize=2.2\hsize}X %
        >{\hsize=0.6\hsize}Y %
        >{\hsize=0.6\hsize}Y
        >{\hsize=0.6\hsize}Y
    } %
        \toprule
        \multirow{2}{*}{\makecell[l]{Methods}}
        & \multicolumn{3}{c}{Success Rate (\%)} \\ 

        \cmidrule(lr){2-4}
        & DH3 & Allegro & Inspire \\

        \midrule
        \multicolumn{4}{r}{\textcolor{gray}{\textit{GraspClutter6D (235 scenes)}}} \\
        
        \midrule
        $\mathcal{D}(\mathcal{R}, \mathcal{O})$-Grasp~\cite{wei2025dro} & 64.7 & 52.1 & 55.9 \\
        AnyDexGrasp~\cite{fang2025anydexgrasp} & 82.5 & 74.1 & 79.3 \\
        AnyDexGrasp${}^*$ & 82.9 & 75.9 & 81.4 \\
        \model (ours) & \underline{\textbf{84.9}} & \underline{\textbf{76.4}} & \underline{\textbf{82.6}} \\
    
        \midrule
        \multicolumn{4}{r}{\textcolor{gray}{\textit{GraspNet-1Billion (90 scenes)}}} \\
        \midrule
        $\mathcal{D}(\mathcal{R}, \mathcal{O})$-Grasp~\cite{wei2025dro} & 66.2 & 53.3 & 57.3 \\
        AnyDexGrasp~\cite{fang2025anydexgrasp} & 85.2 & 80.8 & 84.4 \\
        AnyDexGrasp${}^*$ & 86.6 & 81.2 & 85.0 \\
        \model (ours) & \underline{\textbf{88.9}} & \underline{\textbf{82.2}} & \underline{\textbf{86.9}} \\
        
        \bottomrule
    \end{tabularx}
    \label{tab:sim_vlg_mf}
\end{table}

\begin{table*}[t]
    \centering
    \small
    \caption{\textbf{Experimental results of dynamic grasping on the GraspNet-1Billion~\cite{fang2020graspnet} benchmark.} We report the grasp tracking accuracy (MGTA), mean translation error ($e_{\text{trans}}$), and mean rotation error ($e_{\text{rot}}$) across three difficulty splits (Seen, Similar, and Novel). The best results are highlighted in bold and underlined.}
    \begin{tabular}{lccccccccc}
        \toprule
        \multirow{2}{*}{Methods} &
        \multicolumn{3}{c}{Seen} &
        \multicolumn{3}{c}{Similar} &
        \multicolumn{3}{c}{Novel} \\
        \cmidrule(lr){2-4}\cmidrule(lr){5-7}\cmidrule(lr){8-10}
        & MGTA$\uparrow$ & $e_{\text{trans}}(\mathrm{cm})\downarrow$ & $e_{\text{rot}}(^{\circ})\downarrow$
        & MGTA$\uparrow$ & $e_{\text{trans}}(\mathrm{cm})\downarrow$ & $e_{\text{rot}}(^{\circ})\downarrow$
        & MGTA$\uparrow$ & $e_{\text{trans}}(\mathrm{cm})\downarrow$ & $e_{\text{rot}}(^{\circ})\downarrow$ \\
        \midrule
        Nearest & -0.81 & 15.2 & 95.98 & -0.80 & 15.4 & 97.61 & -0.83 & 16.6 & 96.90 \\
        AnyGrasp~\cite{fang2023anygrasp} & -0.32 & 2.31 & 53.24 & -0.30 & 2.68 & 57.59 & -0.39 & 2.93 & 58.49 \\
        BundleTrack~\cite{wen2021bundletrack} &  0.74 & 1.52 & 15.8  &  0.72 & 1.47 & 16.37 &  0.72 & 1.93 & 15.41 \\
        Target-referenced~\cite{liu2023target} & 0.84 & 1.34 & 10.15 &  0.85 & 1.52 & 9.84  &  0.78 & 1.9  & 12.96 \\
        MotionGrasp~\cite{chen2025motiongrasp} & 0.99 & 1.1 & \underline{\textbf{0.86}} & 0.99 & 1.1 & \underline{\textbf{0.84}} & 0.99 & 1.07 & \underline{\textbf{1.12}} \\

        \midrule
        \model (ours) & \underline{\textbf{1.0}} & \underline{\textbf{0.22}} & 3.57 & \underline{\textbf{1.0}} & \underline{\textbf{0.23}} & 3.68 & \underline{\textbf{1.0}} & \underline{\textbf{0.28}} & 3.28 \\
        \bottomrule
    \end{tabular}
    \label{tab:sim_dynamic_grasping}
\end{table*}

\begin{figure*}[t]
    \centering
    
    \begin{subfigure}{0.47\textwidth}
        \centering
        \includegraphics[width=\textwidth]{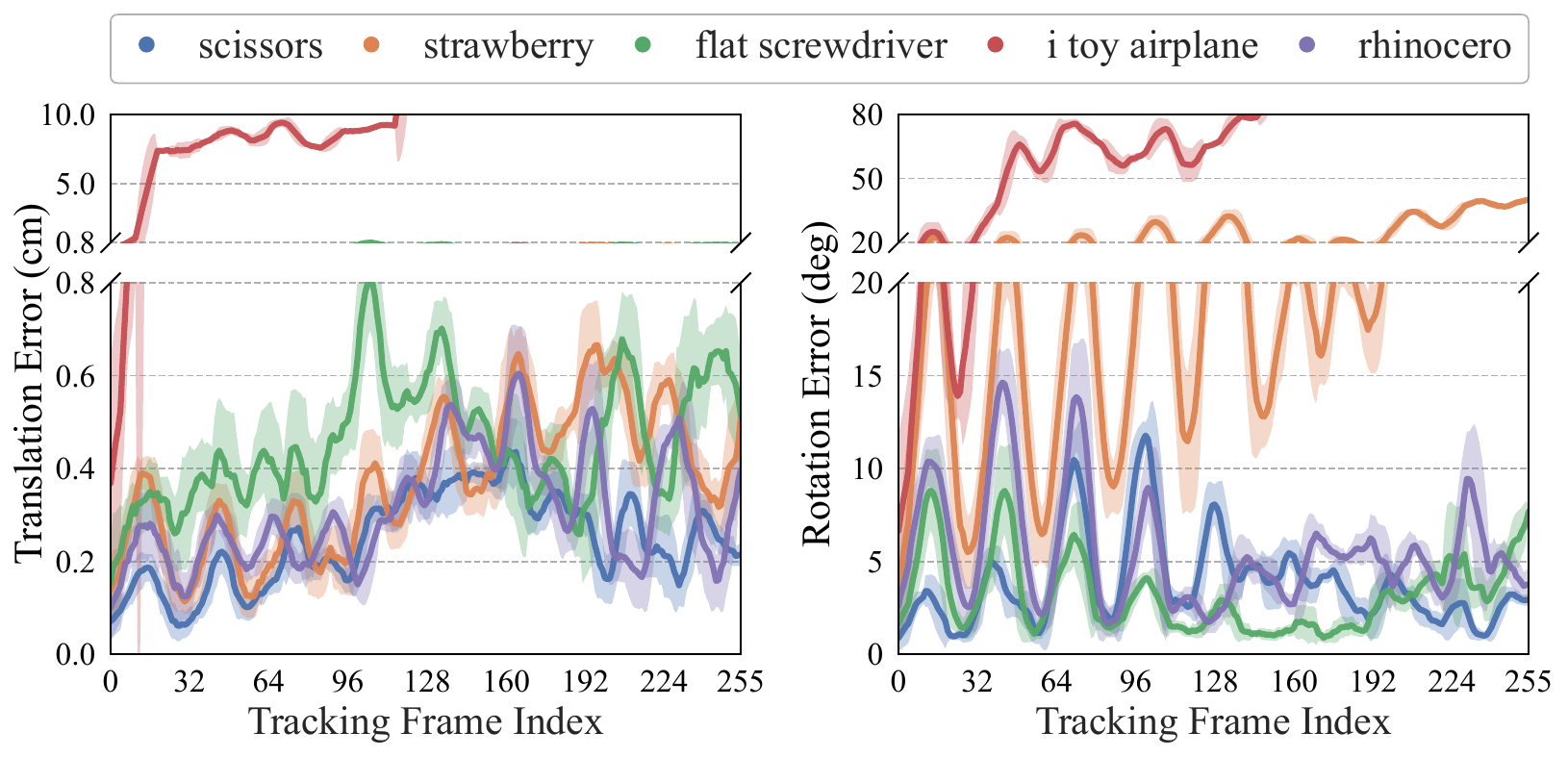}
        \caption{Tracking errors for partial objects in Scene 0100.}
        \label{fig:sim_track_errors}
    \end{subfigure}
    \hfill
    \begin{subfigure}{0.51\textwidth}
        \centering
        \includegraphics[width=\textwidth]{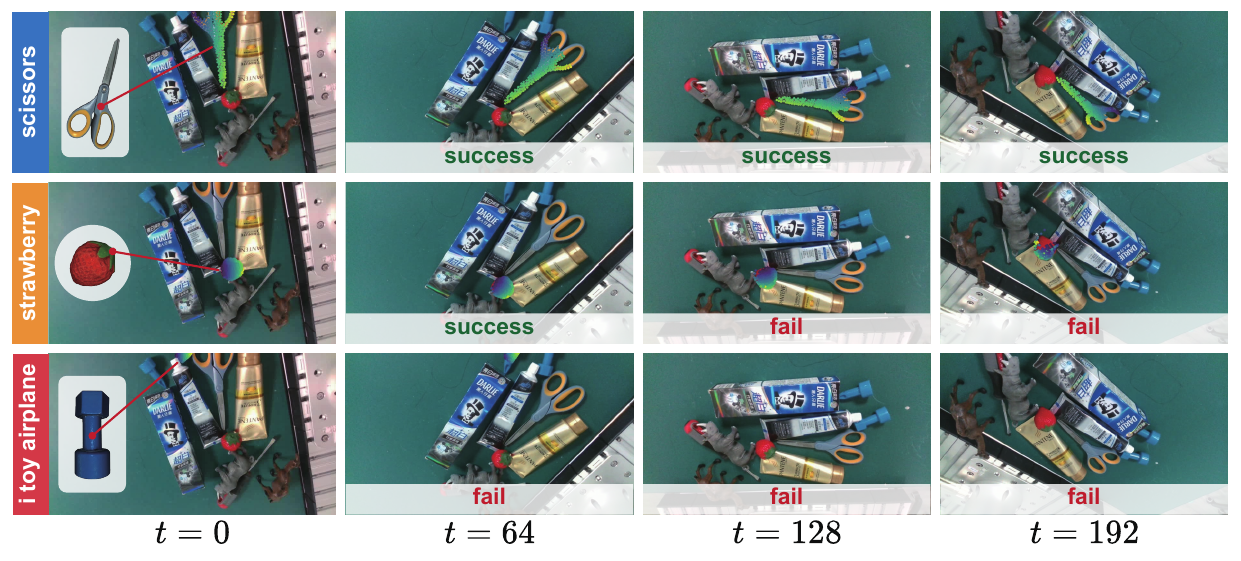}
        \caption{Tracking visualization for partial objects in Scene 0100.}
        \label{fig:sim_track_vis}
    \end{subfigure}
    \caption{\textbf{Long-horizon tracking results for representative objects in scene 0100 of the GraspNet-1Billion~\cite{fang2020graspnet} dataset.} (a) Continuous tracking errors for selected objects. (b) Image-space projections of tracked 3D semantic features.}
    \label{fig:sim_long_horizon_tracking}
\end{figure*}

\subsubsection{Language-Guided Target Grasping}

To evaluate language-guided target grasping, we construct a simulated benchmark in Isaac Sim using the test splits of GraspNet-1Billion~\cite{fang2020graspnet} and GraspClutter6D~\cite{back2025graspclutter6d}. For each scene, objects are initialized with their original poses and assigned a mass of \qty{1.0}{\kilogram} and a friction coefficient of 0.5. The experimental platform consists of a UR5 robotic arm equipped with three different robotic hands: DH3, Allegro, and Inspire. Five static cameras are used to capture the scene, including four surrounding cameras at a $\ang{45}$ elevation angle and one top-down camera. To enable language-guided grasping, we use Qwen3.5-VL\footnote{\url{https://qwen.ai/blog?id=qwen3.5}} to generate language instructions offline by selecting the top five graspable objects in each scene and synthesizing commands following the templates introduced in~\cite{xu2025efficient}. This process produces five predefined language-guided grasping tasks for each scene.

Each task is executed once with each robotic hand. A trial is successful if the hand grasps the target object and lifts it to a height of \qty{10}{\centi\meter}. We report the average grasp \textit{success rate} separately for each hand.

We compare \model with three baseline models: AnyDexGrasp~\cite{fang2025anydexgrasp}, AnyDexGrasp${}^*$, and $\mathcal{D}(\mathcal{R},\mathcal{O})$-Grasp~\cite{wei2025dro}. AnyDexGrasp and AnyDexGrasp${}^*$ have been introduced in \cref{subsubsec:sim_exp_spatial_prior_baselines}. $\mathcal{D}(\mathcal{R},\mathcal{O})$-Grasp is a representative cross-hand dexterous grasping method designed for object-centric grasp synthesis. For fair comparison, all methods are combined with the same cognitive reasoning module as ours for language grounding and target segmentation. AnyDexGrasp and AnyDexGrasp${}^*$ follow the implementation detailed in \cref{subsubsec:sim_exp_spatial_prior_baselines}, utilizing the cognitive reasoning module to filter task-consistent grasp candidates based on the language instruction. For $\mathcal{D}(\mathcal{R},\mathcal{O})$-Grasp, we train the model in a multi-hand setting using grasp samples synthesized via the DexGraspNet~\cite{wang2022dexgraspnet} pipeline on objects from the training sets of both benchmarks. During inference, the cognitive reasoning module first localizes the target object; subsequently, $\mathcal{D}(\mathcal{R},\mathcal{O})$-Grasp generates grasp candidates from the segmented object point cloud and randomly selects a collision-free pose for execution. All methods are evaluated in a zero-shot manner without further fine-tuning, with all grasp types enabled.

As shown in \cref{tab:sim_vlg_mf}, \model achieves the highest success rates on both datasets, outperforming AnyDexGrasp by 2.7\% on GraspClutter6D and 2.5\% on GraspNet-1Billion. Compared with $\mathcal{D}(\mathcal{R}, \mathcal{O})$-Grasp, the average improvements are even more substantial, reaching 23.7\% and 27.1\%, respectively. These results demonstrate that \ac{llm}-based grounding and SAM3-based segmentation provide accurate target constraints, enabling the base policy to generate executable and semantically consistent grasp poses. The combination of the cognitive reasoning module and the base policy yields the best language-guided target grasping performance.

\subsection{Evaluation of Temporal Prior for Dynamic Grasping}
\label{subsec:sim_exp_temporal_prior}
This experiment evaluates the capability of the temporal prior to convert continuous visual observations into temporally consistent motion constraints for grasp updates, thereby enabling robust grasp tracking.

\subsubsection{Benchmark Setting}
We conduct experiments using the GraspNet-1Billion~\cite{fang2020graspnet} benchmark. This benchmark is collected by moving a camera around each scene, which effectively simulates relative object motion and is highly suitable for evaluating dynamic grasping. Each scene contains 256 RGB-D images captured from different viewpoints.

\subsubsection{Evaluation Metrics}
Following prior work~\cite{liu2023target}, we employ \ac{mgta} as the primary performance metric. It measures the alignment between the predicted grasps in the current frame and the ground-truth projections derived from the initial frame. It is defined as:
\begin{equation}
    \label{eq:mgta}
    \text{MGTA} = 1 - \frac{\sum_t (\text{FN}_t + \text{FP}_t + \text{IDSW}_t)}{\sum_t \text{GT}_t},
\end{equation}
where \ac{fn} and \ac{fp} represent missed matches and false associations in the grasp-trajectory association, respectively. \ac{idsw} counts the instances where a grasp is erroneously assigned to a different object ID. An \ac{mgta} score of 1 indicates perfect tracking. Moreover, geometric precision is evaluated using the mean translation error $e_\text{trans}$ and rotation error $e_\text{rot}$ between the predicted grasps and the corresponding ground-truth projections.

\subsubsection{Baseline Models}
We compare \model with five baseline models, comprising Nearest, AnyGrasp~\cite{fang2023anygrasp}, BundleTrack~\cite{wen2021bundletrack}, Target-referenced~\cite{liu2023target}, and MotionGrasp~\cite{chen2025motiongrasp}.

\subsubsection{Experimental Details}
Following the protocol in prior work~\cite{chen2025motiongrasp}, we use the data captured by the RealSense for training and evaluation. To train the baselines, we extract 250 video clips per scene by sequentially grouping 7 consecutive frames. For evaluation, we extract 17 clips from each test scene and use the first 7 frames of each clip for grasp tracking. In the baseline setup, the top-10 highest-scoring grasp candidates are selected from the initial frame for the model to track.

In contrast, our tracking module is entirely training-free. We formulate grasp tracking as a rigid object motion estimation problem. Consequently, geometric precision measurement is equivalent to evaluating the object pose estimation errors. Meanwhile, the \ac{mgta} metric is computed based on instance-level tracking accuracy. For each test clip, we randomly select 2 to 3 objects for tracking. Given their masks in the initial frame, we predict their continuous poses across the video clip and report both the geometric errors and association metrics. In practice, to solve the optimization objective in \cref{eq:prior_d3field_optimization_objective}, we employ the Adam optimizer~\cite{kingma2014adam} for 100 iterations.

\subsubsection{Experimental Results}
As shown in \cref{tab:sim_dynamic_grasping}, our method demonstrates temporally consistent grasp tracking performance in dynamic scenarios. It achieves perfect association accuracy, the lowest translation error, and competitive rotation error. This stable tracking capability is enabled by integrating the robust mask tracking capability of SAM3~\cite{carion2025sam3} with the cross-frame semantic consistency of DINOv3~\cite{simeoni2025dinov3} features.

We further conduct a long-horizon tracking evaluation on representative objects from Scene 0100 of the GraspNet-1Billion~\cite{fang2020graspnet} benchmark. As illustrated in \cref{fig:sim_track_errors}, our method robustly tracks most objects (\eg, the scissors) with average translation and rotation errors around \qty{0.5}{\centi\meter} and $\ang{5}$. However, as shown in \cref{fig:sim_track_vis}, tracking failures primarily emerge in two challenging scenarios: (1) highly symmetric objects (\eg, the strawberry) experience rotation estimation drift due to their rotational invariance, and (2) severely occluded objects with minimal initial visibility (\eg, the i toy airplane) lack sufficient discriminative semantic cues, leading to catastrophic tracking loss and translation deviation during viewpoint changes.

\begin{figure}[t!]
    \centering
    \includegraphics[width=\linewidth]{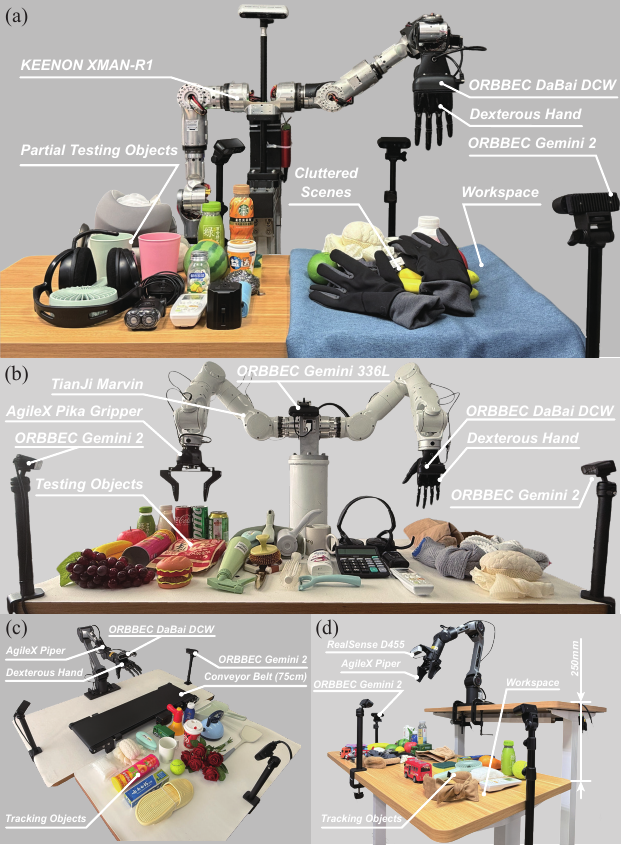}
    \caption{\textbf{Real-world robotic platforms.} We conduct experiments on three robotic platforms with different end-effectors: (a) KEENON XMAN-R1 prototype equipped with a dexterous hand, (b) TianJi Marvin equipped with a dexterous hand and a parallel-jaw gripper, (c) AgileX Piper equipped with a dexterous hand, and (d) AgileX Piper equipped with a parallel-jaw gripper.}
    \label{fig:real_exp_hardware}
\end{figure}

\begin{figure*}[t!]
    \centering
    \includegraphics[width=\linewidth]{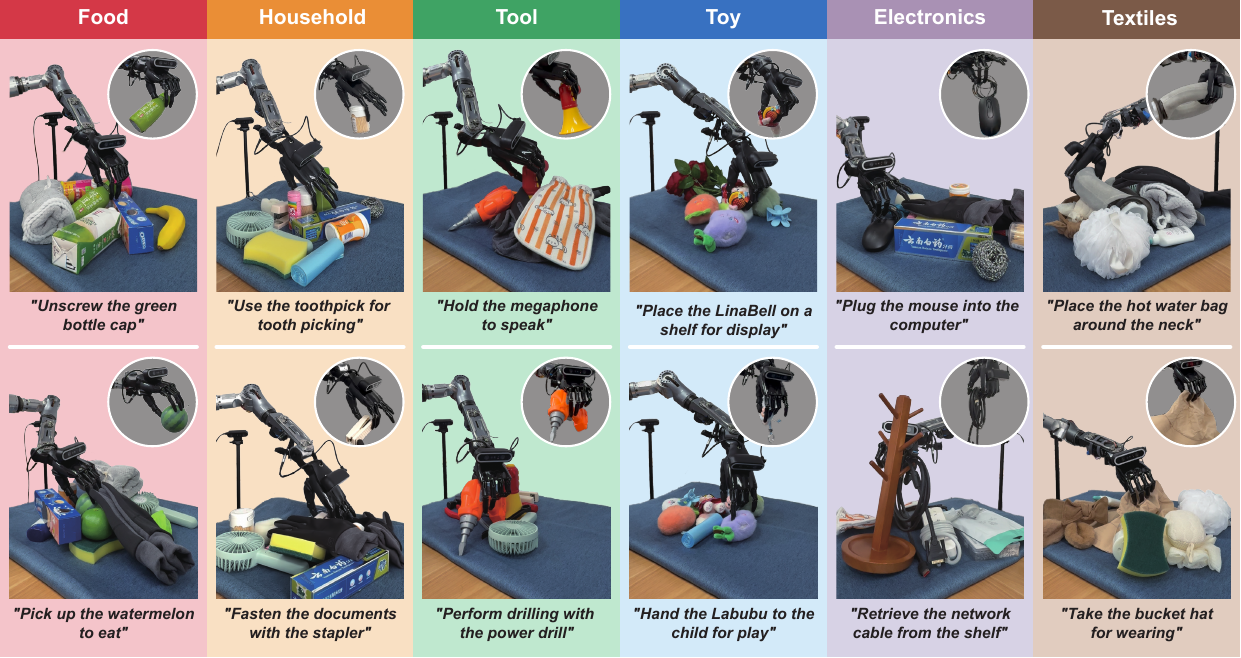}
    \caption{\textbf{Real-world language-guided dexterous functional grasping in static cluttered scenes.} Representative grasping results are presented across six everyday object categories: food, household items, tools, toys, electronics, and textiles.}
    \label{fig:real_functional_grasping}
\end{figure*}

\section{Real-world Experiments}
\label{sec:real_exp}

This section demonstrates the real-world applicability of \model by progressively composing different foundation priors to address increasingly challenging grasping scenarios. After introducing the experimental setup (\cref{subsec:real_exp_setup}), we first evaluate the composition of spatial and cognitive priors for functional grasping in static cluttered scenes (\cref{subsec:real_exp_clutter}). We then further increase task complexity by incorporating the temporal prior to handle functional grasping in dynamic cluttered environments (\cref{subsec:real_exp_dynamic}). Finally, 
we demonstrate the extensibility of \model to new tasks and grasp requirements (\cref{subsec:real_exp_extensibility}).

\subsection{Experimental Setup}
\label{subsec:real_exp_setup}
\subsubsection{Hardware Platforms }
\label{subsubsec:hardware_platforms}
As shown in \cref{fig:real_exp_hardware}, we conduct real-world experiments on three robotic platforms, including the KEENON XMAN-R1 prototype, TianJi Marvin, and AgileX Piper, equipped with two types of end-effectors: a parallel-jaw gripper and a 6-\ac{dof} ROHand dexterous hand. Specifically, on the XMAN-R1 and TianJi Marvin platforms equipped with the dexterous hand, we quantitatively evaluate language-guided functional grasping in cluttered scenes. To further investigate the deployment capability across different end-effectors, we equip the TianJi Marvin with a parallel-jaw gripper and conduct the same evaluation under the identical cluttered-scene setting. Furthermore, on the AgileX Piper platform, we evaluate the system's functional grasping performance in dynamic cluttered scenes and quantitatively verify the temporal robustness of grasp tracking under human-induced disturbances. The real-time inference of the entire system is executed on a workstation equipped with an NVIDIA RTX 3090 GPU.

\subsubsection{Evaluation Protocols}
\label{subsec:real_eval_protocols}
To evaluate the zero-shot performance of our method in the real world, we construct a test set of 102 everyday objects with diverse shapes, materials, and textures, covering six categories: food (22 objects), household items (22), tools (18), toys (13), electronics (13), and textiles (14). A trial is considered successful only when the target object is correctly identified, the executed grasp satisfies the functional requirement, no collision occurs with surrounding objects during execution, and the object is lifted by \SI{10}{\cm} and held for \SI{5}{\s} without being dropped. Under this protocol, we directly evaluate the sim-to-real performance of \model without real-world fine-tuning, using both a parallel-jaw gripper and the ROHand across representative real-world grasping tasks. The feasible grasp types for the ROHand follow the Inspire-hand grasp taxonomy defined in~\cite{fang2025anydexgrasp}.

\subsection{Experiments of Language-Guided Functional Grasping in Static Cluttered Scenes}
\label{subsec:real_exp_clutter}
\subsubsection{Task Description}
In this task, we evaluate the system's ability to perform language-guided functional grasping in dense clutter using the ROHand on both the XMAN-R1 and TianJi Marvin platforms. For each of the 102 target objects, we construct a cluttered scene containing one target object associated with a specific functional instruction, 2--3 distractors with similar semantic categories or visual appearances, and additional background objects. We conduct five trials for each object under randomized object poses and scene configurations, yielding 510 grasping trials in total. We report the category-level and overall success rates aggregated from the two platforms.

\begin{figure}[t!]
    \centering
    \footnotesize
    \begin{subfigure}[t]{0.52\columnwidth}
        \centering
        \includegraphics[width=\linewidth]{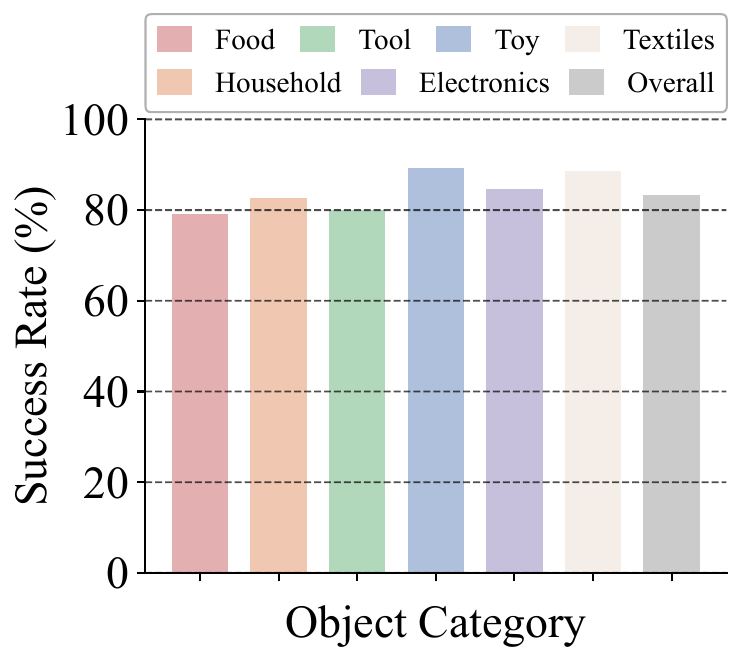}
        \caption{Category success rates.}
        \label{fig:real_functional_success_rates}
    \end{subfigure}\hfill
    \begin{subfigure}[t]{0.44\columnwidth}
        \centering
        \includegraphics[width=\linewidth]{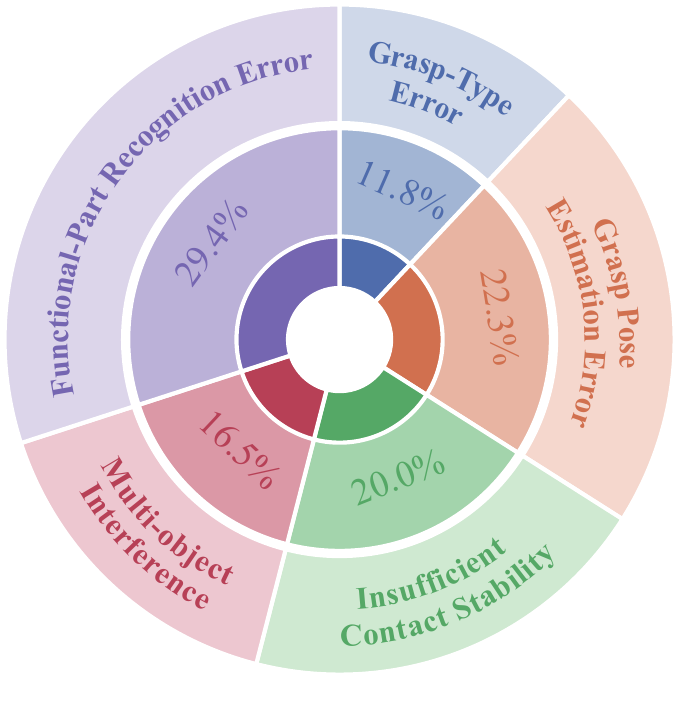}
        \caption{Failure distributions.}
        \label{fig:real_functional_failure_modes}
    \end{subfigure}
    \caption{\textbf{Quantitative analysis for language-guided dexterous functional grasping in cluttered scenes.} The results summarize (a) category-level and overall success rates and (b) failure distributions of unsuccessful trials.}
    \label{fig:real_functional_quantitative_results}
\end{figure}

\subsubsection{Experimental Results}
As shown in \cref{fig:real_functional_grasping}, by composing the spatial and cognitive priors, \model can generate spatially feasible and functionally appropriate dexterous grasp poses in cluttered scenes. When the target object is surrounded by cluttered or occluding objects (\eg, a watermelon and a toothpick box), our method can still reliably plan collision-free grasps. For tools or daily items with explicit functional affordances (\eg, a bottle, a megaphone, and a power drill), it selects task-relevant functional parts and grasp types, producing grasps consistent with human object-use behaviors. As illustrated in \cref{fig:real_functional_success_rates}, \model achieves an overall success rate of 83.3\% across 510 real-world grasping trials. These results demonstrate that the system can reliably perform target identification, functional grasp intent inference, and dexterous grasp generation in real-world clutter.

We further summarize the distribution of failure modes in \cref{fig:real_functional_failure_modes}. Functional-part recognition errors constitute the primary source of failure, accounting for 29.4\% of unsuccessful trials. These failures mainly occur when the target object's functional regions are ambiguous or visually similar to distractors, leading to incorrect functional part localization. Grasp pose estimation errors account for 22.3\%, primarily caused by inaccurate pose estimation or unstable candidate selection under geometric uncertainties. Insufficient contact stability represents 20.0\% of failures, frequently occurring with small or smooth-surfaced objects where the generated grasps fail to maintain reliable contact after lifting. Multi-object interference contributes 16.5\% of failures, where the dexterous hand unintentionally grasps multiple objects in dense clutter. The remaining 11.8\% are attributed to grasp-type errors, which occur when the selected grasp type does not fully match the object's functional requirements.

\begin{figure}[t!]
    \centering
    \includegraphics[width=\linewidth]{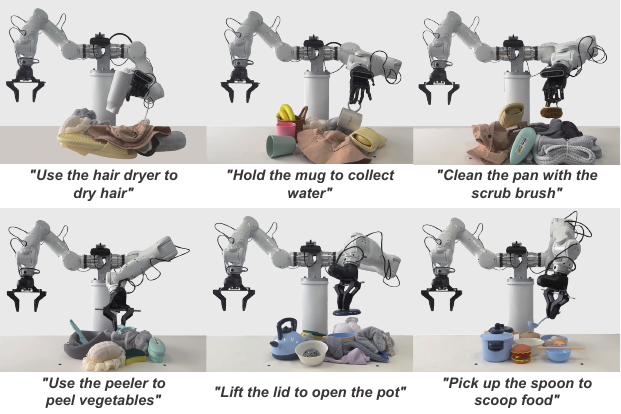}
    \caption{\textbf{Real-world deployment with different end-effectors.} \model performs language-guided functional grasping in cluttered scenes with both a parallel-jaw gripper and the ROHand dexterous hand on the TianJi Marvin platform.}
    \label{fig:real_cross_effector_deployment}
\end{figure}

Additionally, as shown in \cref{fig:real_cross_effector_deployment}, \model performs language-guided functional grasping in cluttered scenes on the TianJi Marvin platform equipped with a parallel-jaw gripper. This evaluation demonstrates that the framework can be readily transferred between different end-effectors.

\begin{figure}[t!]
    \centering
    \includegraphics[width=\linewidth]{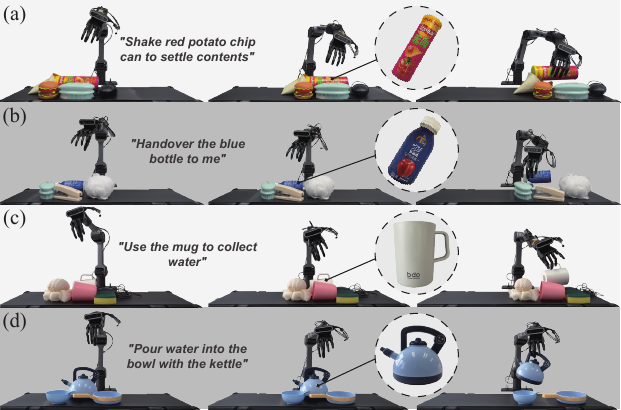}
    \caption{\textbf{Real-world language-guided dexterous functional grasping in dynamic cluttered scenes.} \model performs functional grasping on a moving conveyor belt.}
    \label{fig:real_dynamic_functional_grasping}
\end{figure}

\subsection{Experiments of Language-Guided Functional Grasping in Dynamic Cluttered Scenes}
\label{subsec:real_exp_dynamic}
\subsubsection{Task Description}
This experiment evaluates the ability of the system to perform language-guided functional grasping in dynamic cluttered scenes. It provides a real-world validation of the composability of spatial, cognitive, and temporal priors for handling complex grasping scenarios. The experiment is conducted using the AgileX Piper equipped with a dexterous hand. We select 16 representative objects and place them together with distractors on a moving conveyor belt operating at a speed of \SIrange[per-mode=symbol]{2}{5}{\centi\meter\per\second}. Furthermore, to quantitatively evaluate tracking robustness, we equip the same manipulator with a parallel-jaw gripper. We introduce four types of human-induced disturbances, including out-of-view recovery, fast movement, continuous occlusion, and simultaneous translation and rotation. For each disturbance condition, we select 10 objects and perform five repeated grasping trials, reporting the average success rate across all conditions.

\begin{table}[t!]
    \centering
    \small
    \caption{\textbf{Quantitative evaluation of dynamic grasping.} We report the number of successful trials over total trials under different human-induced disturbances. N/A indicates that the test is not applicable due to rotational symmetry.}
    \resizebox{\columnwidth}{!}{
    \begin{tabular}{lccccc} %
        \toprule
        Object & 
        \makecell{Out-of-view\\Recovery} & 
        \makecell{Fast\\Motion} & 
        \makecell{Continuous\\Occlusion} & 
        \makecell{Trans.\\ \& Rot.} &
        \makecell{Success\\Rate} \\
        \midrule
        Green Bottle    & 5/5 & 4/5 & 5/5 & 4/5 & 90.0\% \\
        Mango           & 5/5 & 4/5 & 5/5 & 4/5 & 90.0\% \\
        Toothpaste      & 5/5 & 4/5 & 5/5 & 4/5 & 90.0\% \\
        Sponge          & 5/5 & 5/5 & 5/5 & 4/5 & 95.0\% \\
        Red Toy Car     & 5/5 & 5/5 & 5/5 & 3/5 & 90.0\% \\
        Tennis Ball     & 5/5 & 4/5 & 5/5 & N/A & 93.3\% \\
        Remote Control  & 5/5 & 5/5 & 4/5 & 4/5 & 90.0\% \\
        Small Fan       & 5/5 & 5/5 & 5/5 & 4/5 & 95.0\% \\
        Hairband        & 4/5 & 4/5 & 4/5 & 3/5 & 75.0\% \\
        White Gauze     & 4/5 & 5/5 & 5/5 & 4/5 & 90.0\% \\
        \midrule
        All             & 48/50 & 45/50 & 48/50 & 34/45 & 89.7\% \\
        \bottomrule
    \end{tabular}
    } %
    \label{tab:dynamic_tracking_robustness}
\end{table}

\begin{figure}[t!]
    \centering
    \includegraphics[width=\linewidth]{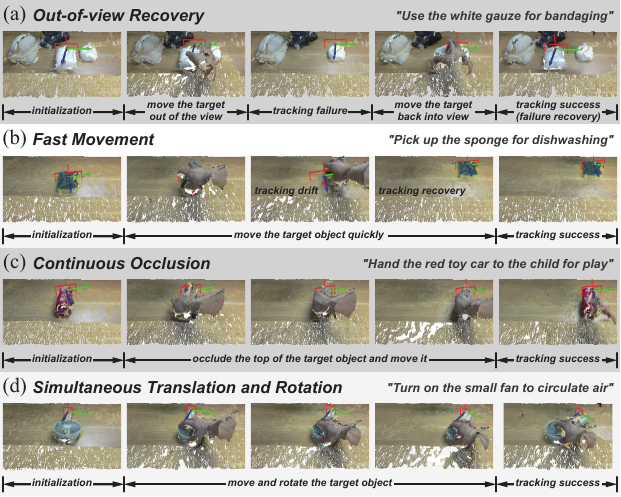}
    \caption{\textbf{Dynamic grasping under different human-induced disturbances.} Each row presents the continuous tracking and grasping process under a specific language instruction.}
    \label{fig:real_disturbance_robustness}
\end{figure}

\subsubsection{Experimental Results}
As illustrated in \cref{fig:real_dynamic_functional_grasping}, our proposed framework demonstrates reliable language-guided functional grasping in dynamic cluttered conveyor-belt scenarios while updating grasp poses online at 5 Hz. For example, in \cref{fig:real_dynamic_functional_grasping}\redtext{c}, given the instruction, the spatial prior constrains grasp generation to avoid collisions with adjacent distractors, while the cognitive prior selects the cup handle as the appropriate functional part and determines Prismatic 3 Finger as the task-suitable grasp type. The base grasp policy then produces stable and executable grasp poses for the dexterous hand. The temporal prior further enables the system to robustly track the target as it moves along the conveyor belt. These results demonstrate the effectiveness of composing the spatial, cognitive, and temporal priors for real-world dynamic grasping.

Furthermore, as reported in \cref{tab:dynamic_tracking_robustness}, the system achieves an average grasping success rate of 89.7\% under various human-induced disturbances (see \cref{fig:real_disturbance_robustness}). Under perturbations such as fast motion, continuous occlusion, simultaneous translation and rotation, the temporal tracking module maintains target feature consistency and continuously updates grasp poses. Notably, even in out-of-view recovery tests, the system rapidly restores object pose estimation and grasp planning, further verifying its robustness in non-static real-world environments.

\begin{figure}[t]
    \centering
    \includegraphics[width=\columnwidth]{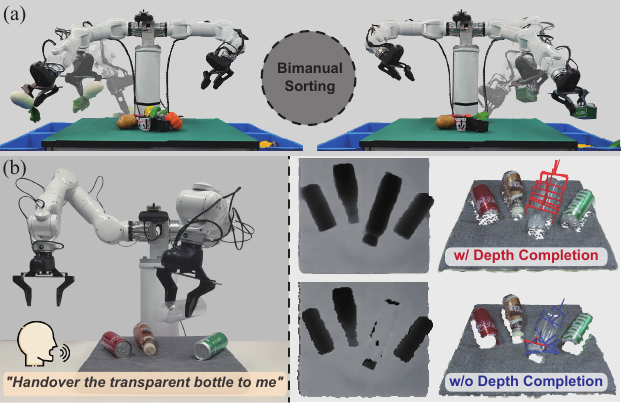}
    \caption{\textbf{Extensibility demonstrations of \model.} We demonstrate two representative extensions: (a) bimanual sorting and (b) transparent-object grasping.}
    \label{fig:real_exp_extensibility}
\end{figure}

\subsection{Extensibility of the Adaptive Framework}
\label{subsec:real_exp_extensibility}

We further demonstrate that our framework can be extended to new challenging grasping scenarios by integrating additional task-relevant modules. As illustrated in \cref{fig:real_exp_extensibility}, incorporating a task-allocation module enables \model to perform bimanual sorting by distributing grasp targets between the two arms. Similarly, transparent-object grasping is supported by integrating a depth-completion module (\eg, SwinDRNet~\cite{dai2022domain}) to recover missing object geometry before constructing the structured grasp interface. This extensibility stems from the decoupling of task understanding and physical grasp synthesis, which allows new capabilities to be incorporated through task-specific modules without requiring an entirely new policy.

\section{Discussion and Conclusion}
\label{sec:discussion_and_conclusion}

\subsection{Limitations and Future Directions}
\label{subsec:limitations_future_directions}

\textbf{Perception under Uncertainty:}
The foundation-model priors for constructing, refining, and updating the structured grasp interface remain sensitive to perceptual uncertainty. Specifically, depth noise can distort contact geometry and compromise grasp generation and evaluation, while inconsistent cross-view grounding can lead to incorrect functional-region identification. Furthermore, frequent changes in visual observations can easily induce pose drift and introduce unnecessary update latency. To address these limitations, future work could integrate more capable foundation models. Promising directions include metric reconstruction~\cite{zhou2026xlens,yang2026robo3r} to improve geometric accuracy, cross-view correspondence~\cite{cao2026vggtdet,pan2026v2sam} to promote grounding consistency, and point-tracking methods~\cite{lin2026point2pose} to support efficient, low-latency pose estimation.

\textbf{Closed-loop Physical Interaction:}
Our base grasp policy predicts grasp stability from visual geometry but lacks tactile feedback to verify contact conditions during execution. Discrepancies between predicted and actual contacts, including insufficient contact forces or incipient slip, may therefore remain undetected, limiting grasp robustness for small or smooth objects. Incorporating tactile feedback~\cite{zhao2025embedding} could complement geometric predictions with direct contact measurements, supporting closed-loop stability assessment and corrective in-hand adjustments.

\textbf{Beyond Single-step Grasping:}
Our framework synthesizes task-consistent grasps but does not explicitly account for dependencies between grasping and other manipulation actions. In heavily cluttered scenes, preparatory actions may be needed to expose the target, while subsequent manipulation may impose additional constraints on grasp selection. Integrating non-prehensile skills~\cite{lin2026dexmove} with sequential task planning could support scene rearrangement before grasping and coordinate grasp selection with subsequent actions.

\subsection{Conclusion}
In this paper, we introduce \model, a task-adaptive \ac{vlg} framework that supports deployment across different robotic hands. It learns an efficient and generalizable base policy for cross-hand grasp synthesis while offloading task-dependent understanding to specialized foundation-model modules. These modules leverage composable priors to convert task contexts into a structured grasp interface, allowing the base policy to generate task-consistent and physically feasible grasp poses. Extensive simulation and real-world experiments demonstrate that \model successfully synthesizes grasps across two- to five-finger end-effectors, while effectively incorporating spatial, cognitive, and temporal priors to adapt to diverse grasping tasks under combined contextual conditions (\eg, cluttered scenes, functional requirements, and dynamic interactions). By establishing a stable interface between high-level task understanding and low-level grasp execution, \model provides a scalable path for challenging grasping scenarios.

{\small
    \bibliographystyle{IEEEtran}
    \bibliography{reference}

@string {CVPR = "Conference on Computer Vision and Pattern Recognition (CVPR)"}

@string {ICCV = "International Conference on Computer Vision (ICCV)"}

@string {ECCV = "European Conference on Computer Vision (ECCV)"}

@string {NeurIPS = "Advances in Neural Information Processing Systems (NeurIPS)"}

@string {ICLR = "International Conference on Learning Representations (ICLR)"}

@string {TMM = "Transactions on Multimedia (TMM)"}

@string {IROS = "International Conference on Intelligent Robots and Systems (IROS)"}

@string {ICRA = "International Conference on Robotics and Automation (ICRA)"}

@string {TRO = "Transactions on Robotics (T-RO)"}

@string {THMS = "Transactions on human-machine systems (THMS)"}

@string {TASE = "Transactions on Automation Science and Engineering (TASE)"}

@string {IJRR = "International Journal of Robotics Research (IJRR)"}

@string {RA-L = "IEEE Robotics and Automation Letters (RA-L)"}

@string {ROMAN = "International Symposium on Robot and Human Interactive Communication (RO-MAN)"}

@string {CHI = "ACM Conference on Human Factors in Computing Systems (CHI)"}

@string {CoRL = "Conference on Robot Learning (CoRL)"}

@article{ravi2024sam2,
  title={SAM 2: Segment Anything in Images and Videos},
  author={Ravi, Nikhila and Gabeur, Valentin and Hu, Yuan-Ting and Hu, Ronghang and Ryali, Chaitanya and Ma, Tengyu and Khedr, Haitham and R{\"a}dle, Roman and Rolland, Chloe and Gustafson, Laura and Mintun, Eric and Pan, Junting and Alwala, Kalyan Vasudev and Carion, Nicolas and Wu, Chao-Yuan and Girshick, Ross and Doll{\'a}r, Piotr and Feichtenhofer, Christoph},
  journal={arXiv preprint arXiv:2408.00714},
  year={2024}
}

@article{simeoni2025dinov3,
  title={Dinov3},
  author={Sim{\'e}oni, Oriane and Vo, Huy V and Seitzer, Maximilian and Baldassarre, Federico and Oquab, Maxime and Jose, Cijo and Khalidov, Vasil and Szafraniec, Marc and Yi, Seungeun and Ramamonjisoa, Micha{\"e}l and others},
  journal={arXiv preprint arXiv:2508.10104},
  year={2025}
}

@inproceedings{sundermeyer2021contact,
  title={Contact-graspnet: Efficient 6-dof grasp generation in cluttered scenes},
  author={Sundermeyer, Martin and Mousavian, Arsalan and Triebel, Rudolph and Fox, Dieter},
  booktitle=ICRA,
  pages={13438--13444},
  year={2021},
}

@inproceedings{fang2020graspnet,
  title={Graspnet-1billion: A large-scale benchmark for general object grasping},
  author={Fang, Hao-Shu and Wang, Chenxi and Gou, Minghao and Lu, Cewu},
  booktitle=CVPR,
  pages={11441--11450},
  year={2020}
}

@article{back2025graspclutter6d,
  title={GraspClutter6D: A Large-scale Real-world Dataset for Robust Perception and Grasping in Cluttered Scenes},
  author={Back, Seunghyeok and Lee, Joosoon and Kim, Kangmin and Rho, Heeseon and Lee, Geonhyup and Kang, Raeyoung and Lee, Sangbeom and Noh, Sangjun and Lee, Youngjin and Lee, Taeyeop and others},
  journal={arXiv preprint arXiv:2504.06866},
  year={2025}
}

@article{shao2020unigrasp,
  title={Unigrasp: Learning a unified model to grasp with multifingered robotic hands},
  author={Shao, Lin and Ferreira, Fabio and Jorda, Mikael and Nambiar, Varun and Luo, Jianlan and Solowjow, Eugen and Ojea, Juan Aparicio and Khatib, Oussama and Bohg, Jeannette},
  journal=RA-L,
  volume={5},
  pages={2286--2293},
  year={2020},
}

@article{wu2025cedex,
  title={CEDex: Cross-Embodiment Dexterous Grasp Generation at Scale from Human-like Contact Representations},
  author={Wu, Zhiyuan and Potamias, Rolandos Alexandros and Zhang, Xuyang and Zhang, Zhongqun and Deng, Jiankang and Luo, Shan},
  journal={arXiv preprint arXiv:2509.24661},
  year={2025}
}

@inproceedings{li2023gendexgrasp,
  title={Gendexgrasp: Generalizable dexterous grasping},
  author={Li, Puhao and Liu, Tengyu and Li, Yuyang and Geng, Yiran and Zhu, Yixin and Yang, Yaodong and Huang, Siyuan},
  booktitle=ICRA,
  pages={8068--8074},
  year={2023},
}

@inproceedings{varley2015generating,
  title={Generating multi-fingered robotic grasps via deep learning},
  author={Varley, Jacob and Weisz, Jonathan and Weiss, Jared and Allen, Peter},
  booktitle=IROS,
  pages={4415--4420},
  year={2015},
}

@article{wu2022learning,
  title={Learning diverse and physically feasible dexterous grasps with generative model and bilevel optimization},
  author={Wu, Albert and Guo, Michelle and Liu, C Karen},
  journal={arXiv preprint arXiv:2207.00195},
  year={2022}
}

@inproceedings{xu2023unidexgrasp,
  title={Unidexgrasp: Universal robotic dexterous grasping via learning diverse proposal generation and goal-conditioned policy},
  author={Xu, Yinzhen and Wan, Weikang and Zhang, Jialiang and Liu, Haoran and Shan, Zikang and Shen, Hao and Wang, Ruicheng and Geng, Haoran and Weng, Yijia and Chen, Jiayi and others},
  booktitle=CVPR,
  pages={4737--4746},
  year={2023}
}

@article{xu2023fast,
  title={Fast force-closure grasp synthesis with learning-based sampling},
  author={Xu, Wei and Guo, Weichao and Shi, Xu and Sheng, Xinjun and Zhu, Xiangyang},
  journal=RA-L,
  volume={8},
  pages={4275--4282},
  year={2023},
}

@inproceedings{wan2023unidexgrasp++,
  title={Unidexgrasp++: Improving dexterous grasping policy learning via geometry-aware curriculum and iterative generalist-specialist learning},
  author={Wan, Weikang and Geng, Haoran and Liu, Yun and Shan, Zikang and Yang, Yaodong and Yi, Li and Wang, He},
  booktitle=ICCV,
  pages={3891--3902},
  year={2023}
}

@inproceedings{xu2021adagrasp,
  title={Adagrasp: Learning an adaptive gripper-aware grasping policy},
  author={Xu, Zhenjia and Qi, Beichun and Agrawal, Shubham and Song, Shuran},
  booktitle=ICRA,
  pages={4620--4626},
  year={2021}
}

@article{chen2025clutterdexgrasp,
  title={ClutterDexGrasp: A Sim-to-Real System for General Dexterous Grasping in Cluttered Scenes},
  author={Chen, Zeyuan and Yan, Qiyang and Chen, Yuanpei and Wu, Tianhao and Zhang, Jiyao and Ding, Zihan and Li, Jinzhou and Yang, Yaodong and Dong, Hao},
  journal={arXiv preprint arXiv:2506.14317},
  year={2025}
}

@article{fang2025anydexgrasp,
  title={AnyDexGrasp: General Dexterous Grasping for Different Hands with Human-level Learning Efficiency},
  author={Fang, Hao-Shu and Yan, Hengxu and Tang, Zhenyu and Fang, Hongjie and Wang, Chenxi and Lu, Cewu},
  journal={arXiv preprint arXiv:2502.16420},
  year={2025}
}

@inproceedings{wei2025dro,
  title={$\mathcal{D}(\mathcal{R}, \mathcal{O})$ grasp: A unified representation of robot and object interaction for cross-embodiment dexterous grasping},
  author={Wei, Zhenyu and Xu, Zhixuan and Guo, Jingxiang and Hou, Yiwen and Gao, Chongkai and Cai, Zhehao and Luo, Jiayu and Shao, Lin},
  booktitle=ICRA,
  pages={4982-4988},
  year={2025},
}

@article{she2022learning,
  title={Learning high-DOF reaching-and-grasping via dynamic representation of gripper-object interaction},
  author={She, Qijin and Hu, Ruizhen and Xu, Juzhan and Liu, Min and Xu, Kai and Huang, Hui},
  journal={arXiv preprint arXiv:2204.13998},
  year={2022}
}

@article{huang2023voxposer,
  title={Voxposer: Composable 3d value maps for robotic manipulation with language models},
  author={Huang, Wenlong and Wang, Chen and Zhang, Ruohan and Li, Yunzhu and Wu, Jiajun and Fei-Fei, Li},
  journal={arXiv preprint arXiv:2307.05973},
  year={2023}
}

@article{yuan2024robopoint,
  title={Robopoint: A vision-language model for spatial affordance prediction for robotics},
  author={Yuan, Wentao and Duan, Jiafei and Blukis, Valts and Pumacay, Wilbert and Krishna, Ranjay and Murali, Adithyavairavan and Mousavian, Arsalan and Fox, Dieter},
  journal={arXiv preprint arXiv:2406.10721},
  year={2024}
}

@article{huang2024rekep,
  title={Rekep: Spatio-temporal reasoning of relational keypoint constraints for robotic manipulation},
  author={Huang, Wenlong and Wang, Chen and Li, Yunzhu and Zhang, Ruohan and Fei-Fei, Li},
  journal={arXiv preprint arXiv:2409.01652},
  year={2024}
}

@article{wang2023d3ield,
  title={D$^3$Fields: Dynamic 3D Descriptor Fields for Zero-Shot Generalizable Rearrangement},
  author={Wang, Yixuan and Zhang, Mingtong and Li, Zhuoran and Kelestemur, Tarik and Driggs-Campbell, Katherine and Wu, Jiajun and Fei-Fei, Li and Li, Yunzhu},
  journal={arXiv preprint arXiv:2309.16118},
  year={2023}
}

@article{ji2024graspsplats,
  title={Graspsplats: Efficient manipulation with 3d feature splatting},
  author={Ji, Mazeyu and Qiu, Ri-Zhao and Zou, Xueyan and Wang, Xiaolong},
  journal={arXiv preprint arXiv:2409.02084},
  year={2024}
}

@article{zheng2024gaussiangrasper,
  title={Gaussiangrasper: 3d language gaussian splatting for open-vocabulary robotic grasping},
  author={Zheng, Yuhang and Chen, Xiangyu and Zheng, Yupeng and Gu, Songen and Yang, Runyi and Jin, Bu and Li, Pengfei and Zhong, Chengliang and Wang, Zengmao and Liu, Lina and others},
  volume={9},
  pages={7827--7834},
  journal=RA-L,
  year={2024},
}

@article{black2024pi_0,
  title={$\pi_0$: A Vision-Language-Action Flow Model for General Robot Control},
  author={Black, Kevin and Brown, Noah and Driess, Danny and Esmail, Adnan and Equi, Michael and Finn, Chelsea and Fusai, Niccolo and Groom, Lachy and Hausman, Karol and Ichter, Brian and others},
  journal={arXiv preprint arXiv:2410.24164},
  year={2024}
}

@article{intelligence2025pi_0.5,
  title={$\pi_{0.5}$: A Vision-Language-Action Model with Open-World Generalization},
  author={Intelligence, Physical and Black, Kevin and Brown, Noah and Darpinian, James and Dhabalia, Karan and Driess, Danny and Esmail, Adnan and Equi, Michael and Finn, Chelsea and Fusai, Niccolo and others},
  journal={arXiv preprint arXiv:2504.16054},
  year={2025}
}

@article{deng2025graspvla,
  title={Graspvla: a grasping foundation model pre-trained on billion-scale synthetic action data},
  author={Deng, Shengliang and Yan, Mi and Wei, Songlin and Ma, Haixin and Yang, Yuxin and Chen, Jiayi and Zhang, Zhiqi and Yang, Taoyu and Zhang, Xuheng and Cui, Heming and others},
  journal={arXiv preprint arXiv:2505.03233},
  year={2025}
}

@article{zhong2025dexgraspvla,
  title={Dexgraspvla: A vision-language-action framework towards general dexterous grasping},
  author={Zhong, Yifan and Huang, Xuchuan and Li, Ruochong and Zhang, Ceyao and Liang, Yitao and Yang, Yaodong and Chen, Yuanpei},
  journal={arXiv preprint arXiv:2502.20900},
  year={2025}
}

@article{xu2025efficient,
  title={Efficient Alignment of Unconditioned Action Prior for Language-conditioned Pick and Place in Clutter},
  author={Xu, Kechun and Xia, Xunlong and Wang, Kaixuan and Yang, Yifei and Mao, Yunxuan and Deng, Bing and Ye, Jieping and Xiong, Rong and Wang, Yue},
  volume={22},
  pages={21256--21268},
  journal=TASE,
  year={2025},
}

@article{fang2023anygrasp,
  title={Anygrasp: Robust and efficient grasp perception in spatial and temporal domains},
  author={Fang, Hao-Shu and Wang, Chenxi and Fang, Hongjie and Gou, Minghao and Liu, Jirong and Yan, Hengxu and Liu, Wenhai and Xie, Yichen and Lu, Cewu},
  volume={39},
  pages={3929--3945},
  journal=TRO,
  year={2023},
}

@article{oquab2023dinov2,
  title={Dinov2: Learning robust visual features without supervision},
  author={Oquab, Maxime and Darcet, Timoth{\'e}e and Moutakanni, Th{\'e}o and Vo, Huy and Szafraniec, Marc and Khalidov, Vasil and Fernandez, Pierre and Haziza, Daniel and Massa, Francisco and El-Nouby, Alaaeldin and others},
  journal={arXiv preprint arXiv:2304.07193},
  year={2023}
}

@article{firoozi2025foundation,
  title={Foundation models in robotics: Applications, challenges, and the future},
  author={Firoozi, Roya and Tucker, Johnathan and Tian, Stephen and Majumdar, Anirudha and Sun, Jiankai and Liu, Weiyu and Zhu, Yuke and Song, Shuran and Kapoor, Ashish and Hausman, Karol and others},
  journal=IJRR,
  volume={44},
  pages={701--739},
  year={2025},
}

@article{bai2025qwen2,
  title={Qwen2. 5-vl technical report},
  author={Bai, Shuai and Chen, Keqin and Liu, Xuejing and Wang, Jialin and Ge, Wenbin and Song, Sibo and Dang, Kai and Wang, Peng and Wang, Shijie and Tang, Jun and others},
  journal={arXiv preprint arXiv:2502.13923},
  year={2025}
}

@article{achiam2023gpt,
  title={Gpt-4 technical report},
  author={Achiam, Josh and Adler, Steven and Agarwal, Sandhini and Ahmad, Lama and Akkaya, Ilge and Aleman, Florencia Leoni and Almeida, Diogo and Altenschmidt, Janko and Altman, Sam and Anadkat, Shyamal and others},
  journal={arXiv preprint arXiv:2303.08774},
  year={2023}
}

@article{RoboBrain2.0,
  title={Robobrain 2.0 technical report},
  author={Team, BAAI RoboBrain and Cao, Mingyu and Tan, Huajie and Ji, Yuheng and Chen, Xiansheng and Lin, Minglan and Li, Zhiyu and Cao, Zhou and Wang, Pengwei and Zhou, Enshen and others},
  journal={arXiv preprint arXiv:2507.02029},
  year={2025}
}

@article{RoboBrain1.0,
    title={Robobrain: A unified brain model for robotic manipulation from abstract to concrete},
    author={Ji, Yuheng and Tan, Huajie and Shi, Jiayu and Hao, Xiaoshuai and Zhang, Yuan and Zhang, Hengyuan and Wang, Pengwei and Zhao, Mengdi and Mu, Yao and An, Pengju and others},
    journal={arXiv preprint arXiv:2502.21257},
    year={2025}
}

@article{yuan2025roboengine,
  title={RoboEngine: Plug-and-Play Robot Data Augmentation with Semantic Robot Segmentation and Background Generation},
  author={Yuan, Chengbo and Joshi, Suraj and Zhu, Shaoting and Su, Hang and Zhao, Hang and Gao, Yang},
  journal={arXiv preprint arXiv:2503.18738},
  year={2025}
}

@article{mei2025rogsam,
  author={Mei, Yunpeng and Sun, Jian and Peng, Zhihong and Deng, Fang and Wang, Gang and Chen, Jie},
  journal=TMM, 
  title={RoG-SAM: A Language-Driven Framework for Instance-Level Robotic Grasping Detection}, 
  year={2025},
  volume={27},
  pages={3057-3068},
}

@article{liu2024rdt,
    title={RDT-1B: a Diffusion Foundation Model for Bimanual Manipulation},
    author={Liu, Songming and Wu, Lingxuan and Li, Bangguo and Tan, Hengkai and Chen, Huayu and Wang, Zhengyi and Xu, Ke and Su, Hang and Zhu, Jun},
    journal={arXiv preprint arXiv:2410.07864},
    year={2024}
}

@article{liu2021synthesizing,
  title={Synthesizing diverse and physically stable grasps with arbitrary hand structures using differentiable force closure estimator},
  author={Liu, Tengyu and Liu, Zeyu and Jiao, Ziyuan and Zhu, Yixin and Zhu, Song-Chun},
  journal=RA-L,
  volume={7},
  pages={470--477},
  year={2021},
}

@article{feix2015grasp,
  title={The grasp taxonomy of human grasp types},
  author={Feix, Thomas and Romero, Javier and Schmiedmayer, Heinz-Bodo and Dollar, Aaron M and Kragic, Danica},
  journal=THMS,
  volume={46},
  pages={66--77},
  year={2015},
}

@inproceedings{dai2015synthesis,
  title={Synthesis and optimization of force closure grasps via sequential semidefinite programming},
  author={Dai, Hongkai and Majumdar, Anirudha and Tedrake, Russ},
  booktitle={Robotics Research},
  volume={2},
  pages={285--305},
  year={2015},
}

@article{carion2025sam3,
  title={Sam 3: Segment anything with concepts},
  author={Carion, Nicolas and Gustafson, Laura and Hu, Yuan-Ting and Debnath, Shoubhik and Hu, Ronghang and Suris, Didac and Ryali, Chaitanya and Alwala, Kalyan Vasudev and Khedr, Haitham and Huang, Andrew and others},
  journal={arXiv preprint arXiv:2511.16719},
  year={2025}
}

@article{wei2022chain,
  title={Chain-of-thought prompting elicits reasoning in large language models},
  author={Wei, Jason and Wang, Xuezhi and Schuurmans, Dale and Bosma, Maarten and Xia, Fei and Chi, Ed and Le, Quoc V and Zhou, Denny and others},
  journal=NeurIPS,
  pages={24824--24837},
  year={2022}
}

@article{lewis2020retrieval,
  title={Retrieval-augmented generation for knowledge-intensive nlp tasks},
  author={Lewis, Patrick and Perez, Ethan and Piktus, Aleksandra and Petroni, Fabio and Karpukhin, Vladimir and Goyal, Naman and K{\"u}ttler, Heinrich and Lewis, Mike and Yih, Wen-tau and Rockt{\"a}schel, Tim and others},
  journal=NeurIPS,
  pages={9459--9474},
  year={2020}
}

@article{chen2025dexonomy,
  title={Dexonomy: Synthesizing All Dexterous Grasp Types in a Grasp Taxonomy},
  author={Chen, Jiayi and Ke, Yubin and Peng, Lin and Wang, He},
  journal={arXiv preprint arXiv:2504.18829},
  year={2025}
}

@article{zhang2025omnidexvlg,
  title={OmniDexVLG: Learning Dexterous Grasp Generation from Vision Language Model-Guided Grasp Semantics, Taxonomy and Functional Affordance},
  author={Zhang, Lei and Zheng, Diwen and Bai, Kaixin and Bing, Zhenshan and Marton, Zoltan-Csaba and Chen, Zhaopeng and Knoll, Alois Christian and Zhang, Jianwei},
  journal={arXiv preprint arXiv:2512.03874},
  year={2025}
}

@article{yan2024dexterous,
  title={Dexterous Manipulation Based on Prior Dexterous Grasp Pose Knowledge},
  author={Yan, Hengxu and Fang, Haoshu and Lu, Cewu},
  journal={arXiv preprint arXiv:2412.15587},
  year={2024}
}

@inproceedings{yan2024surprisingly,
  title={A surprisingly efficient representation for multi-finger grasping},
  author={Yan, Hengxu and Fang, Hao-Shu and Lu, Cewu},
  booktitle=ICRA,
  pages={6462--6469},
  year={2024},
}

@article{santello1998postural,
  title={Postural hand synergies for tool use},
  author={Santello, Marco and Flanders, Martha and Soechting, John F},
  journal={Journal of neuroscience},
  volume={18},
  pages={10105--10115},
  year={1998},
}

@article{mason2001hand,
  title={Hand synergies during reach-to-grasp},
  author={Mason, Carolyn R and Gomez, Jose E and Ebner, Timothy J},
  journal={Journal of neurophysiology},
  volume={86},
  pages={2896--2910},
  year={2001},
}

@article{jarque2016using,
  title={Using kinematic reduction for studying grasping postures. An application to power and precision grasp of cylinders},
  author={Jarque-Bou, N{\'e}stor and Gracia-Ib{\'a}{\~n}ez, Ver{\'o}nica and Sancho-Bru, Joaqu{\'\i}n-Lu{\'\i}s and Vergara, Margarita and P{\'e}rez-Gonz{\'a}lez, Antonio and Andr{\'e}s, FJ},
  journal={Applied ergonomics},
  volume={56},
  pages={52--61},
  year={2016},
}

@article{coumar2025foam,
  title={Foam: A Tool for Spherical Approximation of Robot Geometry},
  author={Coumar, Sai and Chang, Gilbert and Kodkani, Nihar and Kingston, Zachary},
  journal={arXiv preprint arXiv:2503.13704},
  year={2025}
}

@inproceedings{zhang2024dexgraspnet,
  title={DexGraspNet 2.0: Learning Generative Dexterous Grasping in Large-scale Synthetic Cluttered Scenes},
  author={Zhang, Jialiang and Liu, Haoran and Li, Danshi and Yu, XinQiang and Geng, Haoran and Ding, Yufei and Chen, Jiayi and Wang, He},
  booktitle=CoRL,
  pages={5106--5133},
  year={2024}
}

@article{wang2022dexgraspnet,
  title={DexGraspNet: A Large-Scale Robotic Dexterous Grasp Dataset for General Objects Based on Simulation},
  author={Wang, Ruicheng and Zhang, Jialiang and Chen, Jiayi and Xu, Yinzhen and Li, Puhao and Liu, Tengyu and Wang, He},
  journal={arXiv preprint arXiv:2210.02697},
  year={2022}
}

@article{he2025dexvlg,
  title={DexVLG: Dexterous Vision-Language-Grasp Model at Scale},
  author={He, Jiawei and Li, Danshi and Yu, Xinqiang and Qi, Zekun and Zhang, Wenyao and Chen, Jiayi and Zhang, Zhaoxiang and Zhang, Zhizheng and Yi, Li and Wang, He},
  journal={arXiv preprint arXiv:2507.02747},
  year={2025}
}

@inproceedings{wu2024ptv3,
    title={Point Transformer V3: Simpler, Faster, Stronger},
    author={Wu, Xiaoyang and Jiang, Li and Wang, Peng-Shuai and Liu, Zhijian and Liu, Xihui and Qiao, Yu and Ouyang, Wanli and He, Tong and Zhao, Hengshuang},
    booktitle=CVPR,
    pages={4840--4851},
    year={2024}
}

@inproceedings{li2022hgcnet,
  title={HGC-Net: Deep anthropomorphic hand grasping in clutter},
  author={Li, Yiming and Wei, Wei and Li, Daheng and Wang, Peng and Li, Wanyi and Zhong, Jun},
  booktitle=ICRA,
  pages={714--720},
  year={2022},
}

@article{zhang2026cadgrasp,
  title={CADGrasp: Learning Contact and Collision Aware General Dexterous Grasping in Cluttered Scenes},
  author={Zhang, Jiyao and Ma, Zhiyuan and Wu, Tianhao and Chen, Zeyuan and Dong, Hao},
  journal={arXiv preprint arXiv:2601.15039},
  year={2026}
}

@article{chen2025motiongrasp,
  author={Nuo Chen and Xiao{-}Ming Wu and Guohao Xu and Jian{-}Jian Jiang and Zibo Chen and Wei{-}Shi Zheng},
  title={MotionGrasp: Long-Term Grasp Motion Tracking for Dynamic Grasping},
  journal=RA-L,
  volume={10},
  pages={796--803},
  year={2025},
}

@inproceedings{wen2021bundletrack,
  author={Bowen Wen and Kostas E. Bekris},
  title={BundleTrack: 6D Pose Tracking for Novel Objects without Instance or Category-Level 3D Models},
  booktitle=IROS,
  pages={8067--8074},
  year={2021},
}

@inproceedings{liu2023target,
  author={Jirong Liu and Ruo Zhang and Haoshu Fang and Minghao Gou and Hongjie Fang and Chenxi Wang and Sheng Xu and Hengxu Yan and Cewu Lu},
  title={Target-referenced Reactive Grasping for Dynamic Objects},
  booktitle=CVPR,
  pages={8824--8833},
  year={2023},
}

@article{kingma2014adam,
  title={Adam: A method for stochastic optimization},
  author={Kingma, Diederik P and Ba, Jimmy},
  journal={arXiv preprint arXiv:1412.6980},
  year={2014}
}

@article{li2025Language,
  author={Zhuo Li and Junjia Liu and Zhihao Li and Zhipeng Dong and Tao Teng and Yongsheng Ou and Darwin G. Caldwell and Fei Chen},
  title={Language-Guided Dexterous Functional Grasping by {LLM} Generated Grasp Functionality and Synergy for Humanoid Manipulation},
  journal=TASE,
  volume={22},
  pages={10506--10519},
  year={2025},
}

@article{tang2023graspgpt,
  title={Graspgpt: Leveraging semantic knowledge from a large language model for task-oriented grasping},
  author={Tang, Chao and Huang, Dehao and Ge, Wenqi and Liu, Weiyu and Zhang, Hong},
  journal=RA-L,
  volume={8},
  pages={7551--7558},
  year={2023},
}

@article{mirjalili2023lan,
  title={Lan-grasp: Using large language models for semantic object grasping},
  author={Mirjalili, Reihaneh and Krawez, Michael and Silenzi, Simone and Blei, Yannik and Burgard, Wolfram},
  journal={arXiv preprint arXiv:2310.05239},
  year={2023}
}

@article{qian2024thinkgrasp,
  title={Thinkgrasp: A vision-language system for strategic part grasping in clutter},
  author={Qian, Yaoyao and Zhu, Xupeng and Biza, Ondrej and Jiang, Shuo and Zhao, Linfeng and Huang, Haojie and Qi, Yu and Platt, Robert},
  journal={arXiv preprint arXiv:2407.11298},
  year={2024}
}

@article{fei2025trograsp,
  title={T (R, O) Grasp: Efficient Graph Diffusion of Robot-Object Spatial Transformation for Cross-Embodiment Dexterous Grasping},
  author={Fei, Xin and Xu, Zhixuan and Fang, Huaicong and Zhang, Tianrui and Shao, Lin},
  journal={arXiv preprint arXiv:2510.12724},
  year={2025}
}

@inproceedings{song2025robospatial,
  title={Robospatial: Teaching spatial understanding to 2d and 3d vision-language models for robotics},
  author={Song, Chan Hee and Blukis, Valts and Tremblay, Jonathan and Tyree, Stephen and Su, Yu and Birchfield, Stan},
  booktitle=CVPR,
  pages={15768--15780},
  year={2025}
}

@inproceedings{chen2024ll3da,
  title={Ll3da: Visual interactive instruction tuning for omni-3d understanding reasoning and planning},
  author={Chen, Sijin and Chen, Xin and Zhang, Chi and Li, Mingsheng and Yu, Gang and Fei, Hao and Zhu, Hongyuan and Fan, Jiayuan and Chen, Tao},
  booktitle=CVPR,
  pages={26428--26438},
  year={2024}
}

@inproceedings{tang2025affordgrasp,
  title={Affordgrasp: In-context affordance reasoning for open-vocabulary task-oriented grasping in clutter},
  author={Tang, Yingbo and Zhang, Shuaike and Hao, Xiaoshuai and Wang, Pengwei and Wu, Jianlong and Wang, Zhongyuan and Zhang, Shanghang},
  booktitle=IROS,
  pages={9433--9439},
  year={2025},
}

@article{zhao2026affordance,
  title={Affordance-Guided Robotic Grasping via Multimodal Large Language Model Reasoning},
  author={Zhao, Zhou and Gao, Jie and Zheng, Dongyuan},
  journal=TASE,
  year={2026},
  volume={23},
  pages={4088--4100},
}

@inproceedings{liu2025robodexvlm,
  title={Robodexvlm: Visual language model-enabled task planning and motion control for dexterous robot manipulation},
  author={Liu, Haichao and Guo, Sikai and Mai, Pengfei and Cao, Jiahang and Li, Haoang and Ma, Jun},
  booktitle=IROS,
  pages={1381--1388},
  year={2025},
}

@article{meattini2023human,
  author={Meattini, Roberto and Suárez, Raúl and Palli, Gianluca and Melchiorri, Claudio},
  journal=TRO, 
  title={Human to Robot Hand Motion Mapping Methods: Review and Classification}, 
  year={2023},
  volume={39},
  pages={842-861},
}

@article{li2026unidrivevla,
  title={UniDriveVLA: Unifying Understanding, Perception, and Action Planning for Autonomous Driving},
  author={Li, Yongkang and Zhou, Lijun and Yan, Sixu and Liao, Bencheng and Yan, Tianyi and Xiong, Kaixin and Chen, Long and Xie, Hongwei and Wang, Bing and Chen, Guang and others},
  journal={arXiv preprint arXiv:2604.02190},
  year={2026}
}

@article{li2025recogdrive,
  title={ReCogDrive: A Reinforced Cognitive Framework for End-to-End Autonomous Driving},
  author={Li, Yongkang and Xiong, Kaixin and Guo, Xiangyu and Li, Fang and Yan, Sixu and Xu, Gangwei and Zhou, Lijun and Chen, Long and Sun, Haiyang and Wang, Bing and others},
  journal={arXiv preprint arXiv:2506.08052},
  year={2025}
}

@book{thelen1994dynamic,
  title={A dynamic systems approach to the development of cognition and action},
  author={Thelen, Esther and Smith, Linda B},
  year={1994},
  publisher={MIT press}
}

@article{wilson2002six,
  title={Six views of embodied cognition},
  author={Wilson, Margaret},
  journal={Psychonomic bulletin \& review},
  volume={9},
  pages={625--636},
  year={2002},
}

@book{goodale2013sight,
  title={Sight unseen: An exploration of conscious and unconscious vision},
  author={Goodale, Melvyn and Milner, David},
  year={2013},
  publisher={OUP Oxford}
}

@article{zhu2020dark,
  title={Dark, beyond deep: A paradigm shift to cognitive ai with humanlike common sense},
  author={Zhu, Yixin and Gao, Tao and Fan, Lifeng and Huang, Siyuan and Edmonds, Mark and Liu, Hangxin and Gao, Feng and Zhang, Chi and Qi, Siyuan and Wu, Ying Nian and others},
  journal={Engineering},
  volume={6},
  pages={310--345},
  year={2020},
}

@inproceedings{dai2022domain,
  title={Domain randomization-enhanced depth simulation and restoration for perceiving and grasping specular and transparent objects},
  author={Dai, Qiyu and Zhang, Jiyao and Li, Qiwei and Wu, Tianhao and Dong, Hao and Liu, Ziyuan and Tan, Ping and Wang, He},
  booktitle=ECCV,
  pages={374--391},
  year={2022},
}

@article{liu2024reconfigurable,
  title={A reconfigurable data glove for reconstructing physical and virtual grasps},
  author={Liu, Hangxin and Zhang, Zeyu and Jiao, Ziyuan and Zhang, Zhenliang and Li, Minchen and Jiang, Chenfanfu and Zhu, Yixin and Zhu, Song-Chun},
  journal={Engineering},
  volume={32},
  pages={202--216},
  year={2024},
}

@article{beingbeyond2025beingh0,
  title={Being-H0: Vision-Language-Action Pretraining from Large-Scale Human Videos},
  author={Luo, Hao and Feng, Yicheng and Zhang, Wanpeng and Zheng, Sipeng and Wang, Ye and Yuan, Haoqi and Liu, Jiazheng and Xu, Chaoyi and Jin, Qin and Lu, Zongqing},
  journal={arXiv preprint arXiv:2507.15597},
  year={2025}
}

@article{choi2026autodex,
  title  = {AutoDex: An Automated Real-World System for Dexterous Grasping Data Collection},
  author = {Choi, Mingi and Kim, Gunhee and Kim, Jisoo and Kim, Taeksoo and Ha, Taeyun and Lim, Jongbin and Joo, Hanbyul},
  journal={arXiv preprint arXiv:2606.23689},
  year={2026}
}

@article{tang2025foundationgrasp,
  title={Foundationgrasp: Generalizable task-oriented grasping with foundation models},
  author={Tang, Chao and Huang, Dehao and Dong, Wenlong and Xu, Ruinian and Zhang, Hong},
  journal=TASE,
  year={2025},
  volume={22},
  pages={12418-12435},
}

@article{cao2026vggtdet,
  title={Vggt-det: Mining vggt internal priors for sensor-geometry-free multi-view indoor 3d object detection},
  author={Cao, Yang and Wu, Feize and Chen, Dave Zhenyu and Zhong, Yingji and Hong, Lanqing and Xu, Dan},
  journal={arXiv preprint arXiv:2603.00912},
  year={2026}
}

@inproceedings{pan2026v2sam,
  title={V\^{} 2-SAM: Marrying SAM2 with Multi-Prompt Experts for Cross-View Object Correspondence},
  author={Pan, Jiancheng and Wang, Runze and Qian, Tianwen and Mahdi, Mohammad and Fu, Yanwei and Xue, Xiangyang and Huang, Xiaomeng and Van Gool, Luc and Paudel, Danda Pani and Fu, Yuqian},
  booktitle=CVPR,
  pages={16910--16919},
  year={2026}
}

@article{yang2026robo3r,
  title={Robo3R: Enhancing Robotic Manipulation with Accurate Feed-Forward 3D Reconstruction},
  author={Yang, Sizhe and Xu, Linning and Li, Hao and Mu, Juncheng and Zeng, Jia and Lin, Dahua and Pang, Jiangmiao},
  journal={arXiv preprint arXiv:2602.10101},
  year={2026}
}

@article{zhou2026xlens,
  title={X-Lens: Real-Time Metric Depth Estimation with Heterogeneous Cameras},
  author={Zhou, Heng and Liu, Shuhong and He, Yonghao and Zhang, Bohao and Fu, Fa and Hou, Chenhui and Hou, Xianbao and Han, Lijun and Sui, Wei},
  journal={arXiv preprint arXiv:2607.12993},
  year={2026}
}

@article{zhao2025embedding,
    author = {Zhao, Zihang and Li, Wanlin and Li, Yuyang and Liu, Tengyu and Li, Boren and Wang, Meng and Du, Kai and Liu, Hangxin and Zhu, Yixin and Wang, Qining and others},
    journal = {Nature Machine Intelligence},
    title = {Embedding high-resolution touch across robotic hands enables adaptive human-like grasping},
    volume = {7},
    pages = {889--900},
    year = {2025},
}

@inproceedings{lin2026dexmove,
  title={DexMove: Learning Tactile-Guided Non-Prehensile Manipulation with Dexterous Hands},
  author={Pei, Lin and Yuzhe, Huang and Wanlin, Li and Chenxi, Xiao and Ziyuan, Jiao},
  booktitle=ICLR,
  year={2026},
}

@article{lin2026point2pose,
  title={Point2Pose: Occlusion-Recovering 6D Pose Tracking and 3D Reconstruction for Multiple Unknown Objects Via 2D Point Trackers},
  author={Lin, Tzu-Yuan and Lee, Ho Jae and Doherty, Kevin and Lee, Yonghyeon and Kim, Sangbae},
  journal={arXiv preprint arXiv:2604.10415},
  year={2026}
}
}





\end{document}